%% file: main.tex
\documentclass[10pt,twocolumn,letterpaper]{article}

\usepackage[accsupp]{axessibility} 

\usepackage[final]{cvpr}      
\usepackage{wrapfig}
\usepackage[dvipsnames,table]{xcolor}
\definecolor{lightblue}{rgb}{0.678, 0.804, 0.82}
\definecolor{lightgray}{rgb}{0.7, 0.75, 0.8}
\usepackage{arydshln}

\input{preamble}

\definecolor{cvprblue}{rgb}{0.21,0.49,0.74}
\usepackage{algpseudocode}

\usepackage[pagebackref,breaklinks,colorlinks,citecolor=cvprblue]{hyperref}

\input{math_commands.tex}
\usepackage[ruled, vlined]{algorithm2e}
\usepackage{xr}

\def\paperID{12014} 
\def\confName{CVPR}
\def\confYear{2024}

\title{Mosaic-SDF for 3D Generative Models}

\author{Lior Yariv$^{1,3}$ \thanks{Work done while interning at Meta.}
\and \hspace{20pt}
Omri Puny$^{3}$
\and \hspace{20pt}
Oran Gafni$^{1}$
\and \hspace{20pt}
Yaron Lipman$^{2,3}$
\vspace{0pt}
\and 
$^{1}$GenAI, Meta
\and $^{2}$FAIR, Meta
\and $^{3}$Weizmann Institute of Science
}

\begin{document}
\maketitle
\input{sec/0_abstract}    
\input{sec/1_intro}
\input{sec/2_relatedwork}
\input{sec/3_method}
\input{sec/4_experiments}

\section{Summary and Future Work}
We presented a novel 3D shape representation, Mosaic-SDF, that is geared towards 3D generative models and offers a simple and efficient preprocessing, favorable approximation-parameter trade-off, and a simple tensorial structure compatible with powerful modern architectures (\ie, transformers). We have used M-SDF to train Flow Matching generative models and demonstrated state of the art results for forward-based models. We believe that M-SDF is the method of choice for 3D generation however still posses some limitations and can be extended in several ways: First, currently we only encode the SDF, missing texture/color/light information. An interesting extension is to incorporate texture and/or light field data. Second, in our architecture we use a simple linear layer passing the local grids into the transformer. A possible extension here is to incorporate convolution layers and/or autoencoders to further increase resolution/data reuse of the representation. Lastly, making M-SDF equivariant to orientations, \eg, by adding local coordinate frames, can improve the generalization abilities of the trained model, which is currently only permutation equivariant.

\section*{Acknowledgements}\vspace{-3pt}
We thank Itai Gat, Matt Le, Ricky Chen for valuable advice and discussion, and Biao Zhang for sharing evaluation code.
OP is supported by a grant from Israel CHE Program for Data Science Research Centers. 


{
    \small
    \bibliographystyle{ieeenat_fullname}
    \bibliography{main}
}

\input{sec/X_suppl}

\end{document}

%% file: preamble.tex
\usepackage[dvipsnames]{xcolor}


%% file: math_commands.tex
\usepackage{xcolor}
\definecolor{lightgray}{gray}{0.9}

\newcommand{\norm}[1]{\left\Vert#1\right\Vert}

\newcommand{\set}[1]{\left\{#1\right\}}
\newcommand{\parr}[1]{\left (#1\right )}
\newcommand{\brac}[1]{\left [#1\right ]}

\newcommand{\Real}{\mathbb R}

\newcommand{\too}{\rightarrow}

\newcommand{\dist}{\textrm{d}} 

\definecolor{mygray}{gray}{0.95}

\usepackage{amsmath,amsfonts,bm}

\def\eqref#1{equation~\ref{#1}}

\def\1{\bm{1}}

\def\rvx{{\mathbf{x}}}
\def\rvy{{\mathbf{y}}}

\def\vc{{\bm{c}}}

\def\vp{{\bm{p}}}

\def\vx{{\bm{x}}}
\def\vy{{\bm{y}}}

\def\mP{{\bm{P}}}

\def\mV{{\bm{V}}}

\DeclareMathAlphabet{\mathsfit}{\encodingdefault}{\sfdefault}{m}{sl}
\SetMathAlphabet{\mathsfit}{bold}{\encodingdefault}{\sfdefault}{bx}{n}

\def\gB{{\mathcal{B}}}

\def\gD{{\mathcal{D}}}

\def\gG{{\mathcal{G}}}

\def\gL{{\mathcal{L}}}

\def\gN{{\mathcal{N}}}

\def\gR{{\mathcal{R}}}
\def\gS{{\mathcal{S}}}

\def\gU{{\mathcal{U}}}

\def\gX{{\mathcal{X}}}
\def\gY{{\mathcal{Y}}}

\def\sB{{\mathbb{B}}}

\def\sG{{\mathbb{G}}}

\newcommand{\E}{\mathbb{E}}

\DeclareMathOperator*{\argmin}{arg\,min}

\usepackage{tabularx,ragged2e,booktabs,caption}
\newcolumntype{C}[1]{>{\Centering}m{#1}}

\newcolumntype{Z}[1]{>{\Left}m{#1}}

\usepackage{pifont}

%% file: sec/0_abstract.tex
\begin{abstract}
Current diffusion or flow-based generative models for 3D shapes divide to two: distilling pre-trained 2D image diffusion models, and training directly on 3D shapes. When training a diffusion or flow models on 3D shapes a crucial design choice is the shape representation. An effective shape representation needs to adhere three design principles: it should allow an efficient conversion of large 3D datasets to the representation form; it should provide a good tradeoff of approximation power versus number of parameters; and it should have a simple tensorial form that is compatible with existing powerful neural architectures. While standard 3D shape representations such as volumetric grids and point clouds do not adhere to all these principles simultaneously, we advocate in this paper a new representation that does. We introduce Mosaic-SDF (M-SDF): a simple 3D shape representation that approximates the Signed Distance Function (SDF) of a given shape by using a set of local grids spread near the shape's boundary. The M-SDF representation is fast to compute for each shape individually making it readily parallelizable; it is parameter efficient as it only covers the space around the shape's boundary; and it has a simple matrix form, compatible with Transformer-based architectures. 
We demonstrate the efficacy of the M-SDF representation by using it to train a 3D generative flow model including class-conditioned generation with the ShapeNetCore-V2 (3D Warehouse) dataset, and text-to-3D generation using a dataset of about 600k caption-shape pairs.


\end{abstract}

%% file: sec/1_intro.tex
\section{Introduction}
\label{sec:intro}

Image-based generative models have rapidly advanced in recent years due to improvements in generation methodologies (\eg, Diffusion Models), model architecture and conditioning (\eg, Text-to-Image Models, Attention/Transformer layers), and the consolidation of large, high quality image datasets. Although generation of 3D shapes have progressed as well, it has not seen the same level of progress demonstrated in image generation. 

Current works for 3D generation divide mostly into two groups: \emph{Optimization based}: 2D image generative priors are used for training/optimizing 3D shape \cite{poole2022dreamfusion,lin2023magic3d,wang2023prolificdreamer}. The main benefit is leveraging existing powerful image models, but the generation process is expensive/slow as it requires training a new model for each sample. Furthermore, using only image priors is often insufficient to build consistent 3D shape (\eg, the Janus effect) \cite{wiki:janus}. \emph{Forward based}: the 3D shapes are generated by a forward process, such as solving an Ordinary Differnetial Euqation (ODE), making generation process more efficient than optimization based methods. However, forward based model are first trained on a dataset of 3D shapes using, \eg, Diffusion or Flow Models, where to that end shapes are first transformed into some canonical 3D representation, \eg, volumetric grid or a point cloud. Forward based models work directly with 3D shapes and suffer from two limitations: first, they require 3D shape datasets for training and these still lag behind image datasets in terms of quantity and quality. Second, in contrast to images, 3D shapes do not occupy the full 3D space and therefore an effective 3D shape representation is more challenging to find/work with.

In this work we focus on forward based 3D generative models. To enable high quality, large scale 3D generation the 3D shape representation of choice should adhere the following design principles: 
\begin{enumerate}[label=(\roman*)]
    \item \textbf{Preprocess efficiency}: can be efficiently computed for a large collection of shapes.
    \item \textbf{Parameter efficiency}: provides a good approximation vs.~parameter count trade-off. 
    \item \textbf{Simple structure}: has a simple tensorial structure, compatible  with expressive neural architectures.
\end{enumerate}
Examining existing 3D representations used for training generative 3D models including volumetric grids \cite{muller2023diffrf, li2023diffusionsdf}, tri-planes \cite{shue20223d,gupta20233dgen,Wang_2023_CVPR}, point clouds \cite{luo2021diffusion, zeng2022lion, Zhou_2021_ICCV, nichol2022pointe}, meshes \cite{Liu2023MeshDiffusion} and neural fields \cite{jun2023shape} we find that all are lacking in one or more of the above design principles. For example, volumetric grids scale cubically and maintain redundant information in the entire space, while tri-planes tend to be more efficient but still encode full 3D tensors and require a neural network trained jointly on the training set. Point clouds do not provide a full surface information; meshes do not enjoy a simple tensorial structure in general; and neural fields are encoded by weight vector with non-trivial symmetries \cite{navon2023equivariant}.

The goal of this work is to introduce Mosaic-SDF (M-SDF), a simple novel 3D shape representation that achieves all desired properties listed above. With M-SDF we are able to train a forward based flow generative model on a datasets of 3D shapes achieving high quality and diversity. In a nutshell, M-SDF approximates an arbitrary Signed Distance Function (SDF) with a \emph{set} of small ($7\times 7\times 7$) volumetric grids with  different centers and scales. Namely, M-SDF of a single shape is a matrix $X$ of dimension $n\times d$, where each row represent a single grid, and it can be fitted to a given shape's SDF in less then 2 minutes with a single Nvidia A100 GPU. Furthermore, as $X$ is a matrix representing a set it is compatible with existing expressive architectures, similar to the ones trained on points clouds \cite{nichol2022pointe}. Lastly, since the grids are centered near the shape's surface M-SDF provides a parameter efficient representation. 

We have used the M-SDF representation to train a Flow Matching model \cite{lipman2023flow} on two main datasets: ShapeNetCore-V2 (3D Warehouse) \cite{shapenet2015} consisting of $\sim$50K 3D models divided to 55 classes and a dataset of $\sim$600K shapes and a matching text description \cite{luo2023scalable}.  We find that M-SDF to be the method of choice for 3D geometry representation and facilitate simple, generalizable, high quality, forward based 3D generative models.

\begin{figure}[t]
    \centering
    \includegraphics[width=\columnwidth]{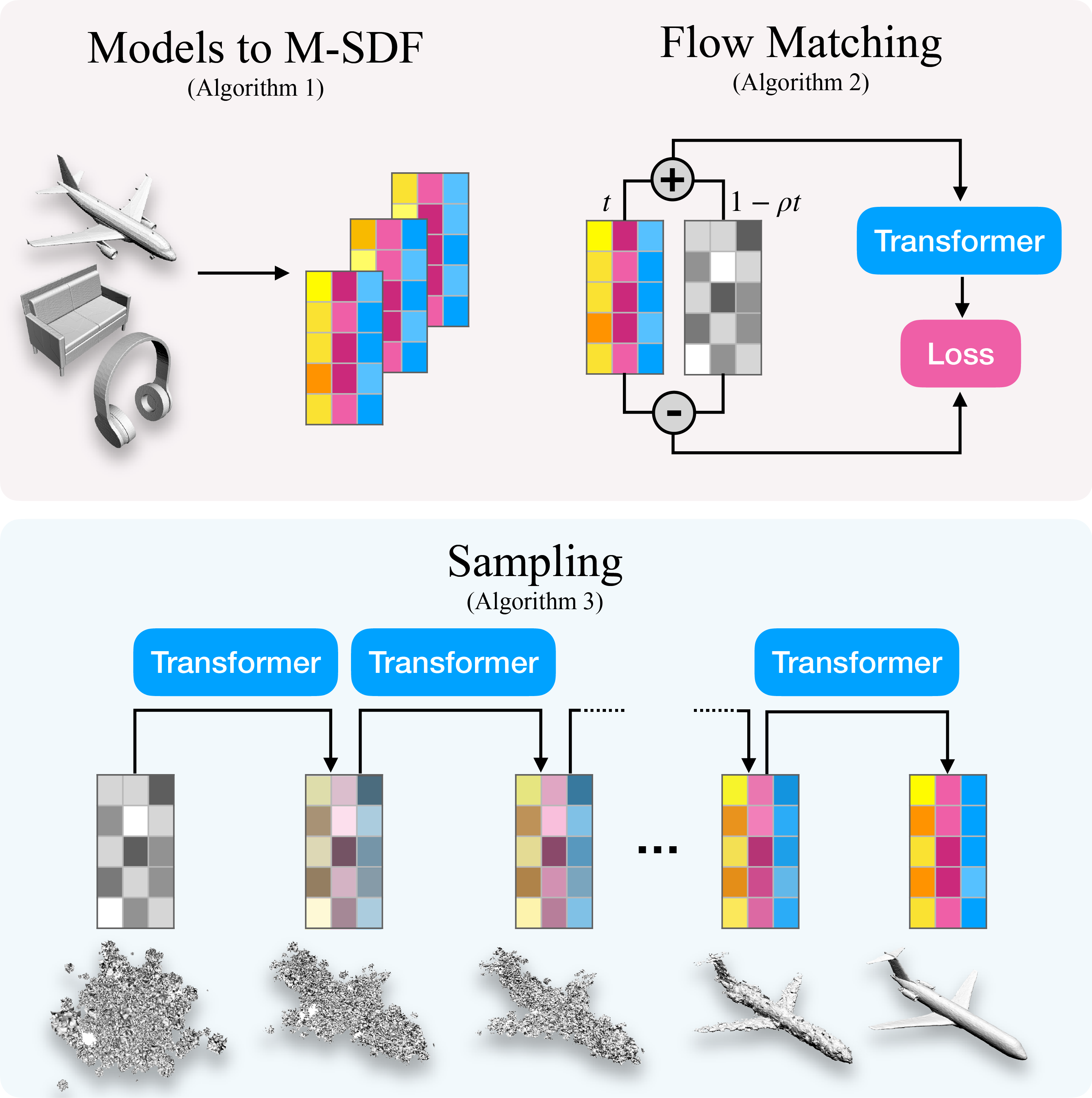}
    \caption{Method overview. Train (top): First we convert the dataset of shapes to M-SDF representations (Algorithm \ref{alg:mosaicsdf}), next we train a Flow Matching model with the M-SDF representations (Algorithm \ref{alg:fm_training}). Sampling (bottom): We random a noisy M-SDF and numerically solve the ode in \eqref{e:ode} (Algorithm \ref{alg:sampling}). }
    \label{fig:enter-label}
\end{figure}

%% file: sec/2_relatedwork.tex
\section{Related work}
\label{sec:relatedwork}
Our work focuses on designing a representation that is beneficial for 3D generation. Below we review relevant previous works categorized by the type of representation they used for 3D shapes. 


\vspace{3pt}
\textbf{Grids and Triplanes.}
Several works suggested to train a diffusion model to predict a grid-based representation, which can be either 3D voxel grids \cite{muller2023diffrf, li2023diffusionsdf, hui2022wavelet}, or using the Triplanes representation \cite{shue20223d,gupta20233dgen,Wang_2023_CVPR}.
%
%
DiffRF \cite{muller2023diffrf} represents each shape as a volumetric radiance field on a voxel grid. \cite{li2023diffusionsdf} represents voxelized truncated signed distance functions (TSDF) encoded into a lower resolution features grid. Neural Wavelet \cite{hui2022wavelet} also advocates voxel grid structure containing wavelet coefficients of a truncated SDF signal. 
%
The main drawback of voxel grids is that they are restricted to a fixed and usually low resolution grids, mainly due to their cubic scaling. 


The cubic scaling of volumetric grids motivated the Triplane representation \cite{Chan2021, kplanes_2023} using lower dimensional grids (1D and 2D) and encode a 3D function as a small MLP applied to sums of outer products. \cite{shue20223d,gupta20233dgen,Wang_2023_CVPR} utilize 2D diffusion model backbones on 3D training scenes represented by 2D feature planes.  
Despite the elegant correspondence to 2D images, prepossessing a large data set of 3D shapes into Triplane representation is compute intensive, requires to learn a shared MLP and in some cases a shared auto-encoder \cite{gupta20233dgen, Wang_2023_CVPR}. 
Compared to our representation, grid-based representations contains many redundant empty voxels, since the shape's surface usually occupied only a small fraction of the grid.

\vspace{3pt}
\textbf{Neural Fields.}
Following the considerable success of Implicit Neural Representations (INRs) for modeling 3D shapes \cite{Park_2019_CVPR,OccupancyNetworks,mildenhall2020nerf}, several works suggest a generative model that produces a parameterization of an implicit function.
\cite{chen2018implicit_decoder,zheng2022sdfstylegan} employed a GAN to generate latent vector or volume representing an implicit function with a shared MLP.
Shape-E \cite{jun2023shape} suggests a diffusion model that directly predicts the weights of a Multi-Layer Perceptron (MLP).
However, training an MLP for each shape in a large dataset is compute intensive, and should consider the symmetries of MLP weights \cite{navon2023equivariant}.


\vspace{3pt}
\textbf{Point clouds.}
\cite{luo2021diffusion, zeng2022lion, Zhou_2021_ICCV, nichol2022pointe} suggest to generate 3D Point clouds, taking advantage of existing permutation-equivariant architectures \cite{qi2016pointnet, NIPS2017_3f5ee243}. Although point-clouds are easy to compute during preprocess and hence suitable for training on large datasets, post-processing it into a smooth surface results in loss of details or requires extensive number of points, which are hard to generate with permutation-equivariant models \cite{wagstaff2022universal}. Point-E \cite{nichol2022pointe} suggests to train an additional upsampler diffusion model to scale the size of the generated point cloud, however it still fails to describe thin structures.



\vspace{3pt}
\textbf{Other representations.}
There are papers that suggested methods that are related to our work and not fall into one of the categories above. \cite{mbr_gvp} offers to generate a set of Volumetric Primitives \cite{Lombardi21} composed on top a coarse mesh, supervised with another 3D generative model's outputs. \cite{zhang2022dilg} and \cite{zhang20233dshape2vecset} presented different representation for encoding occupancy field using a set of either structured or unstructured latent vectors.
Similarly to our representation, they are compatible with the transformer architecture \cite{NIPS2017_3f5ee243}, trained as the generative model.
However, all these methods require a significant compute intensive preprocessing stage for fitting a generalized representation for a large scale dataset. In contrast, our representation is both compact, has a simple structure, interpretable, and can be formed independently and quickly for each shape in the dataset.




%% file: sec/3_method.tex
\section{Method}
\label{sec:method}

In this section we present our 3D shape representation based on an efficient approximation of the Signed Distance Function (SDF), we detail how it is computed for a dataset of shapes in a preprocess stage, and how it is used for training a flow-based generative model. 

\subsection{Mosaic-SDF Shape Representation}
We advocate a simple and effective representation for the Signed Distance Functions (SDFs) of 3D shapes, suitable for generative models, that we call \emph{Mosaic-SDF} (M-SDF). Our main focus is on building a representation that satisfies properties (i)-(iii) from Section \ref{sec:intro}. \vspace{-10pt}

\paragraph{Signed Distance Function (SDF).}
Our goal is to build a simple low parameter count approximation to the SDF of a shape $\gS\subset \Real^3$, where $\vx\in\gS$ are points \emph{inside} the shape, $\vx\notin \gS$ are \emph{outside} and $x\in\partial \gS$ are on the shape's \emph{boundary}. The SDF of the shape is defined by
\begin{equation}
    F_\gS(\vx) = \begin{cases}
    -\dist(\vx,\partial \gS) & \vx\in \gS \\
    \dist(\vx,\partial \gS) & \vx \notin \gS 
    \end{cases},
\end{equation}
where the unsigned distance function is defined by $\dist(\vx,\partial \gS)=\min_{\vy\in\partial\gS}\norm{\vx-\vy}_2$.\vspace{-10pt}

\begin{wrapfigure}[8]{r}{0.38\columnwidth}
  \begin{center}\vspace{-35pt}\hspace{-15pt}
    \includegraphics[width=0.4\columnwidth]{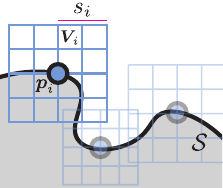}
  \end{center}\vspace{-20pt}
  \caption{Mosaic-SDF representation.} \label{fig:msdf_illustration}
\end{wrapfigure}
\paragraph{M-SDF representation.} We approximate $F_\gS$ with a volumetric function $F_X:\Real^3\too\Real$ defined by the parameters $X$. The core idea is to define $F_X$ as a weighted combination of a \emph{set of local grids}. Following the three principles outlined above, (i)-(iii) we define the representation $X$ to be the set of tuples
\begin{equation}
    X = \set{ (\vp_i, s_i, \mV_i)}_{i\in [n]},
\end{equation}
where $[n]=\set{1,\ldots,n}$, $\vp_i \in \Real^3$ are 3D point locations, $s_i \in \Real$ are local scales, and $\mV_i\in\Real^{k\times k\times k}$ are local volumetric grids. See Figure \ref{fig:msdf_illustration} for an illustration of the local grids. We denote by $I_{\mV_i}:[-1,1]^3\too\Real$ the trilinear interpolants of the values $\mV_i$ over the origin-centered volumetric grid $\sG$ of the cube $[-1,1]^3$ defined by 
\begin{equation}\label{e:G}
    \sG = \set{\frac{2\parr{i_1,i_2,i_3}-n-1}{n-1}\Big\vert i_1,i_2,i_3\in [k]}.
\end{equation}
By convention the interpolants $I_{\mV_i}$ vanish outside the cube $[-1,1]^3$, \ie, $I_{\mV_i}(\vx)=0$ for $\vx\notin [-1,1]^3$. The parametric SDF approximation is then defined by 
\begin{equation}\label{e:F_X}
    F_X(\vx) = \sum_{i\in [n]} w_i\parr{\vx} I_{\mV_i}\parr{\frac{\vx-\vp_i}{s_i}}
\end{equation}
where $w_i(\cdot)$ are scalar weight functions that define the contribution of the $i$'th local grid; the $w_i(\cdot)$ are supported in the cube $[-1,1]^3$ and satisfying partition of unity, \ie, 
\begin{equation}
 \sum_{i \in [n]} w_i(\vx)=1, \qquad \forall \vx \in \Real^3. 
\end{equation}
We opt for 
\begin{equation*}
    w_i(\vx) = \frac{\bar{w}_i(\vx)}{\sum_{j\in [n]}\bar{w}_j(\vx)}, \bar{w}_i(\vx) = {\scriptstyle \mathrm{ReLU}}\hspace{-2pt}\brac{1\hspace{-2pt} - \hspace{-2pt}\norm{\frac{\vx - \vp_i}{s_i}}_\infty}.
\end{equation*}
The domain of definition of $F_X$ is the union of local grids,
\begin{equation}\label{e:D}
    \gD(X) = \cup_{i\in[n]}\sB_\infty(\vp_i,s_i),
\end{equation} 
where $\sB_\infty(\vp,s)=\set{\vx\in\Real^3\vert \norm{\vx-\vp}_\infty<s}$ is the infinity ball of radius $s$ centered at $\vp$.

\begin{figure*}
\centering
\begin{tabular}{@{\hspace{1pt}}c@{\hspace{1pt}}c@{\hspace{1pt}}c@{\hspace{1pt}}c@{\hspace{1pt}}c}    
\includegraphics[width=0.2\textwidth]{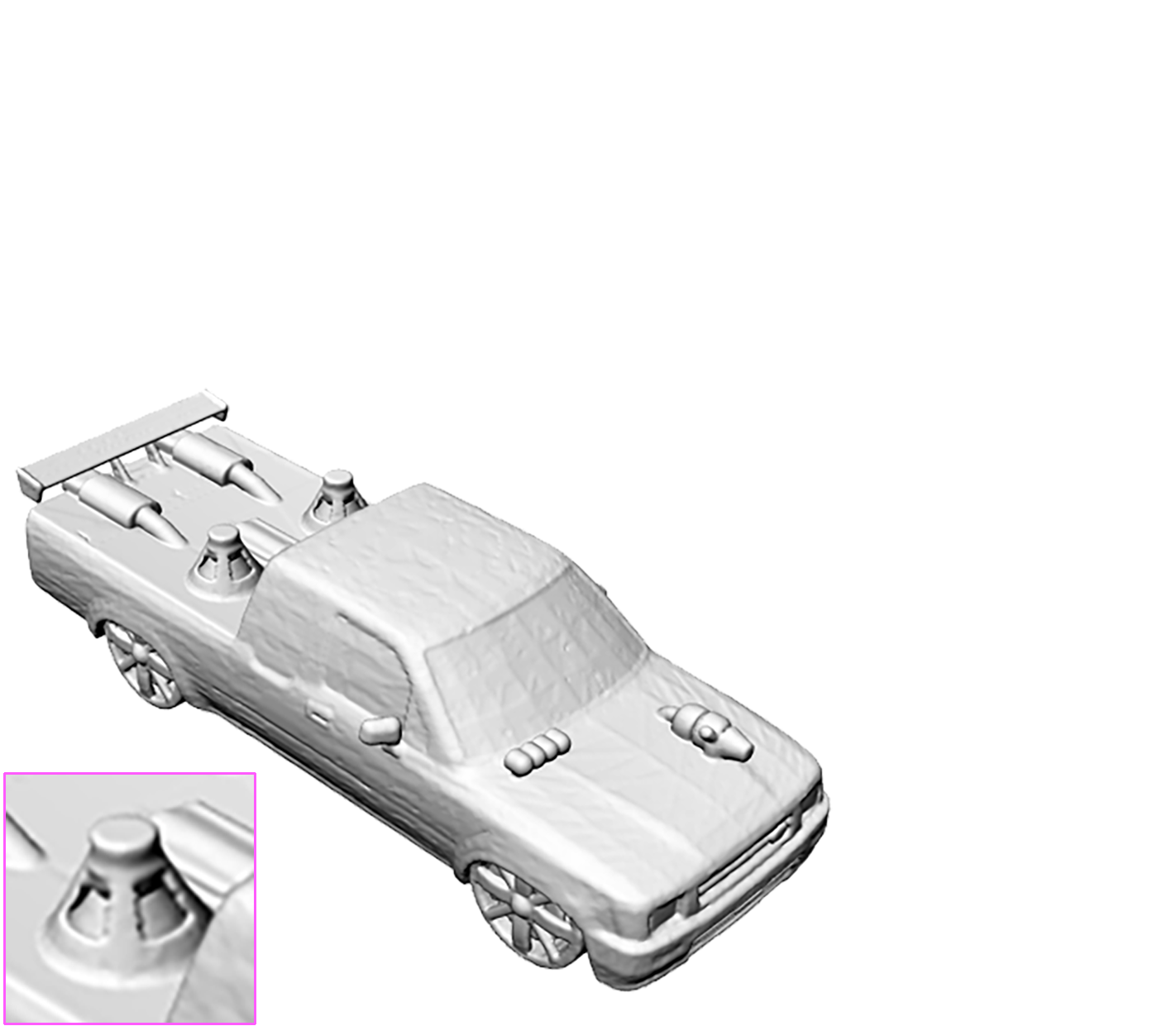} & 
\includegraphics[width=0.2\textwidth]{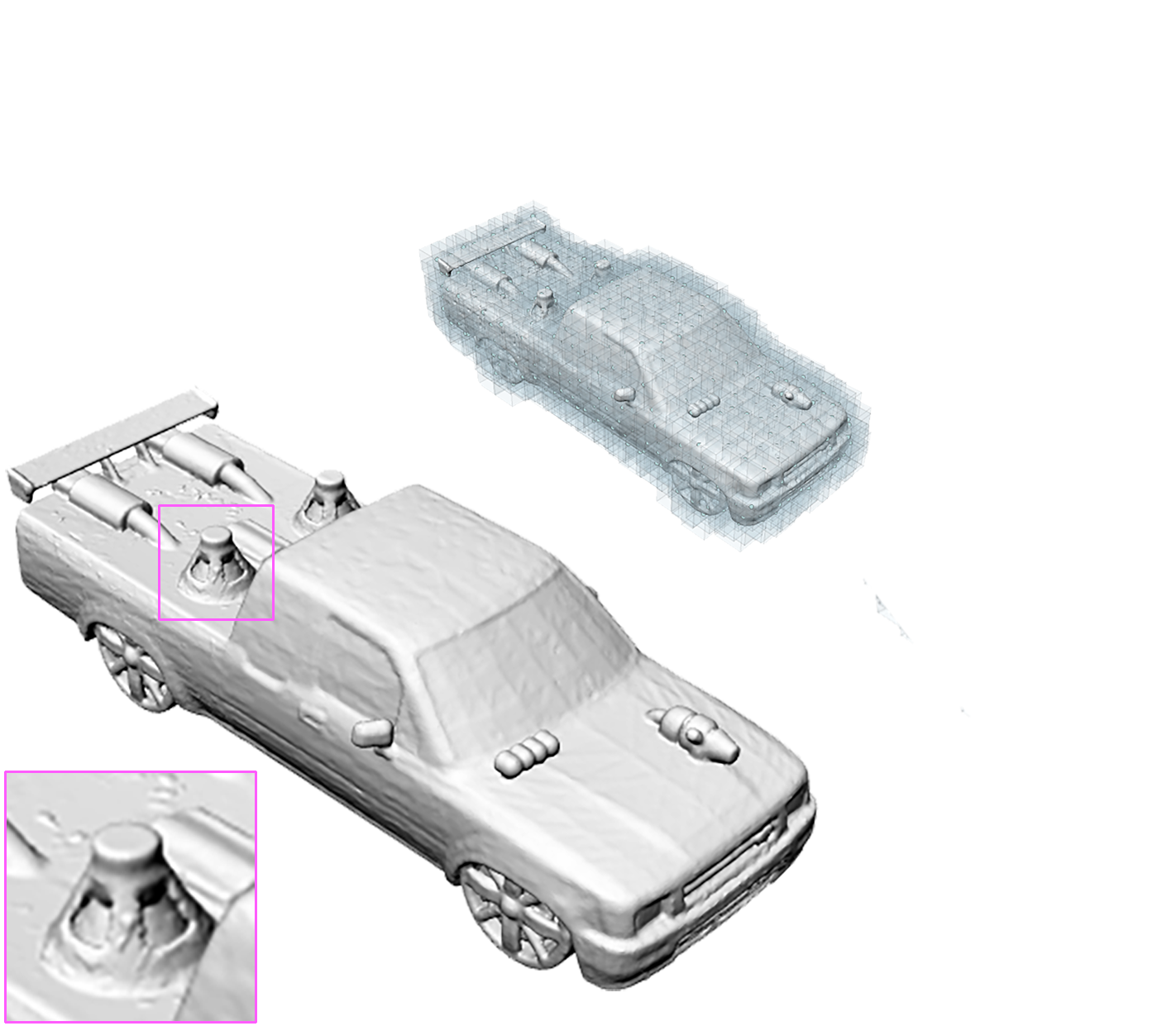} & 
\includegraphics[width=0.2\textwidth]{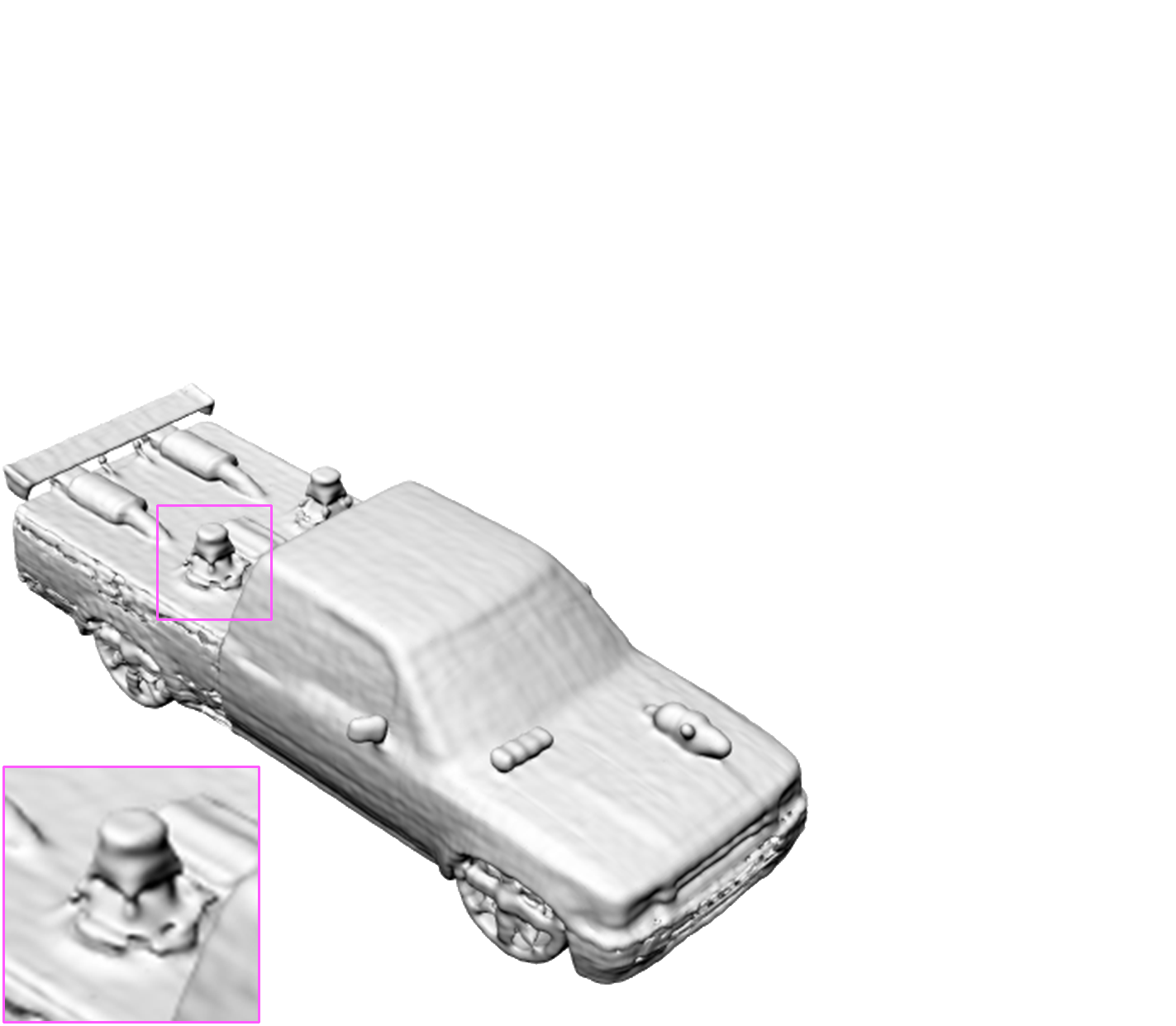} & 
\includegraphics[width=0.2\textwidth]{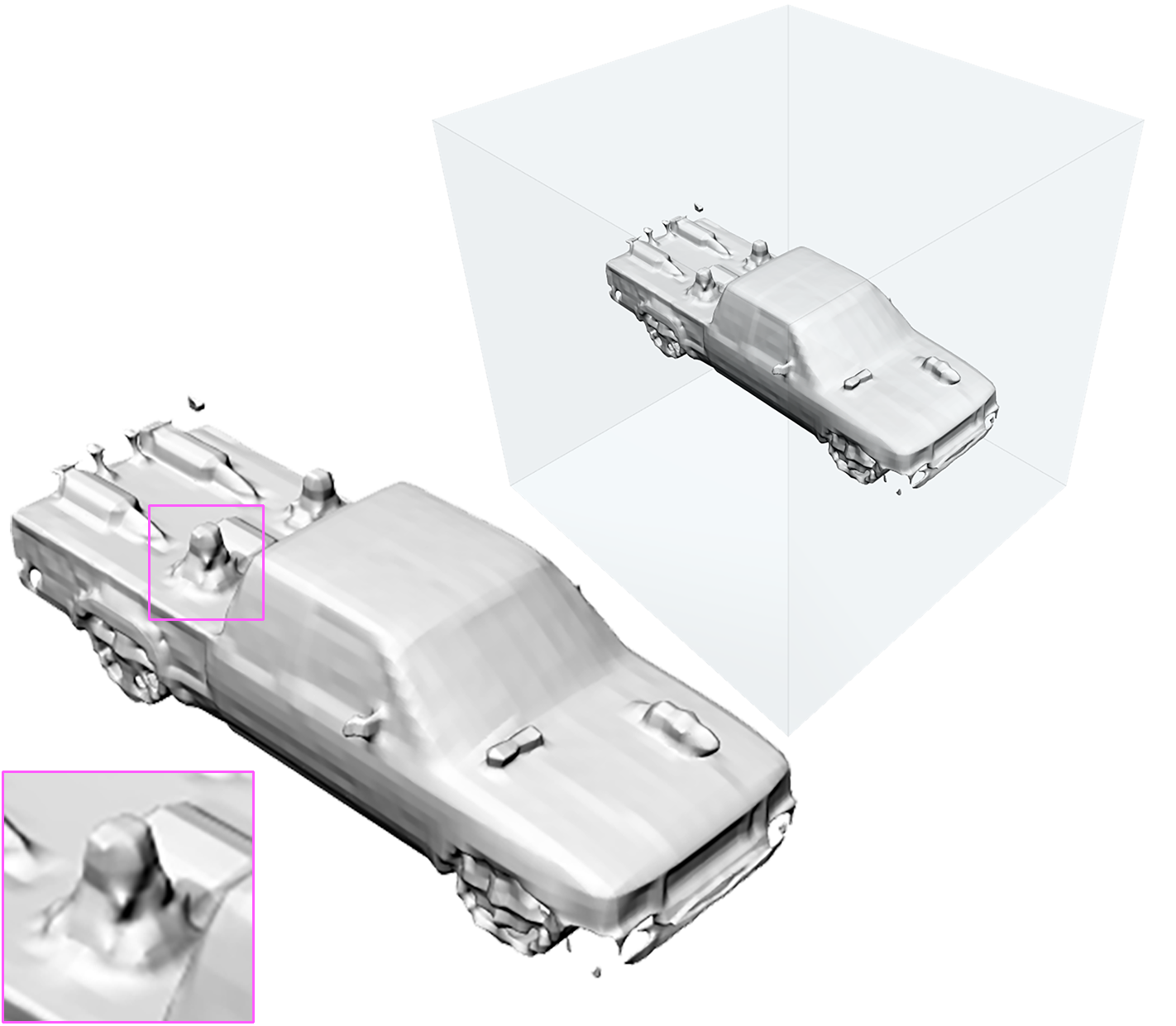} & 
\includegraphics[width=0.2\textwidth]{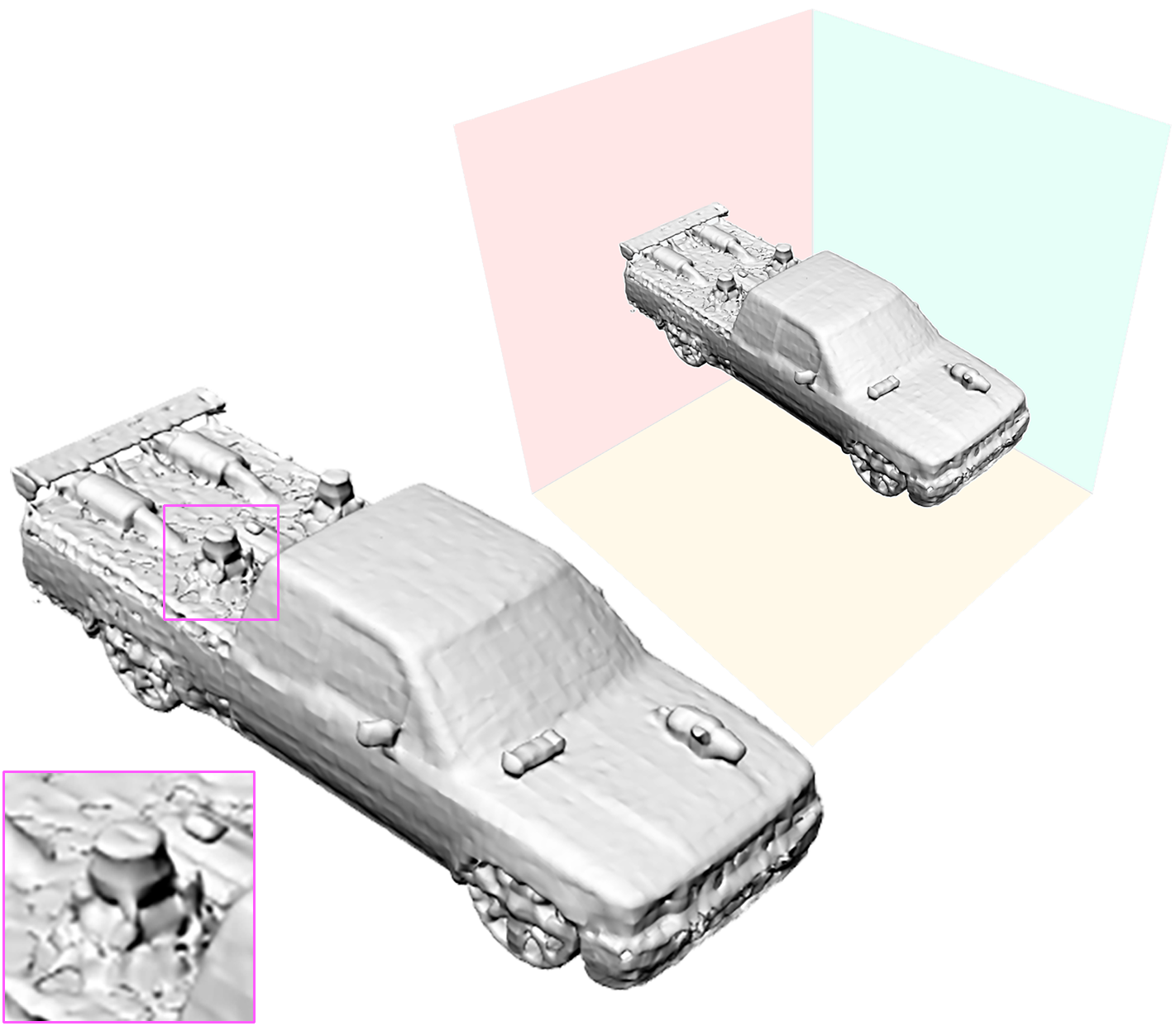} \\
\includegraphics[width=0.2\textwidth]{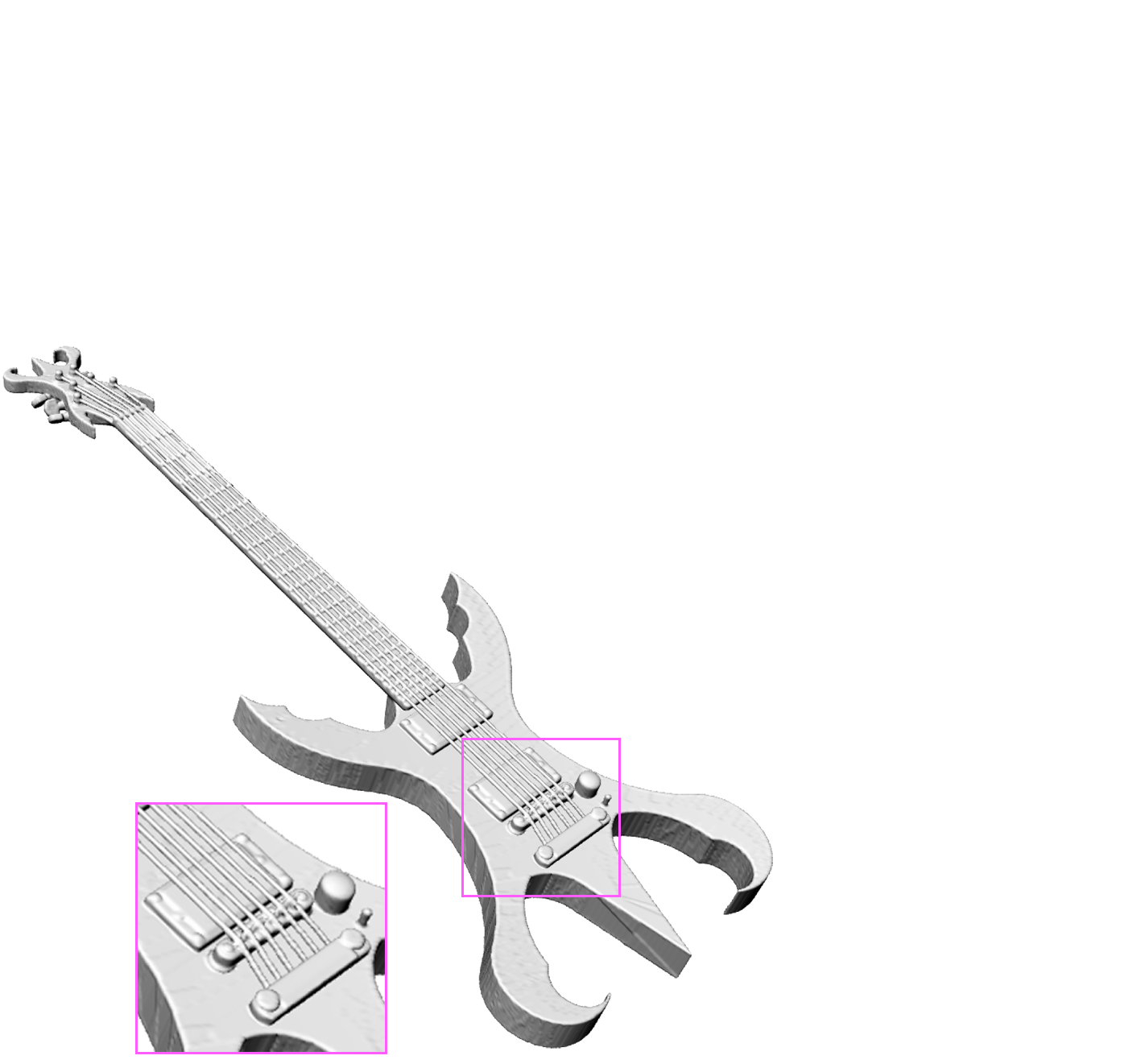} & 
\includegraphics[width=0.2\textwidth]{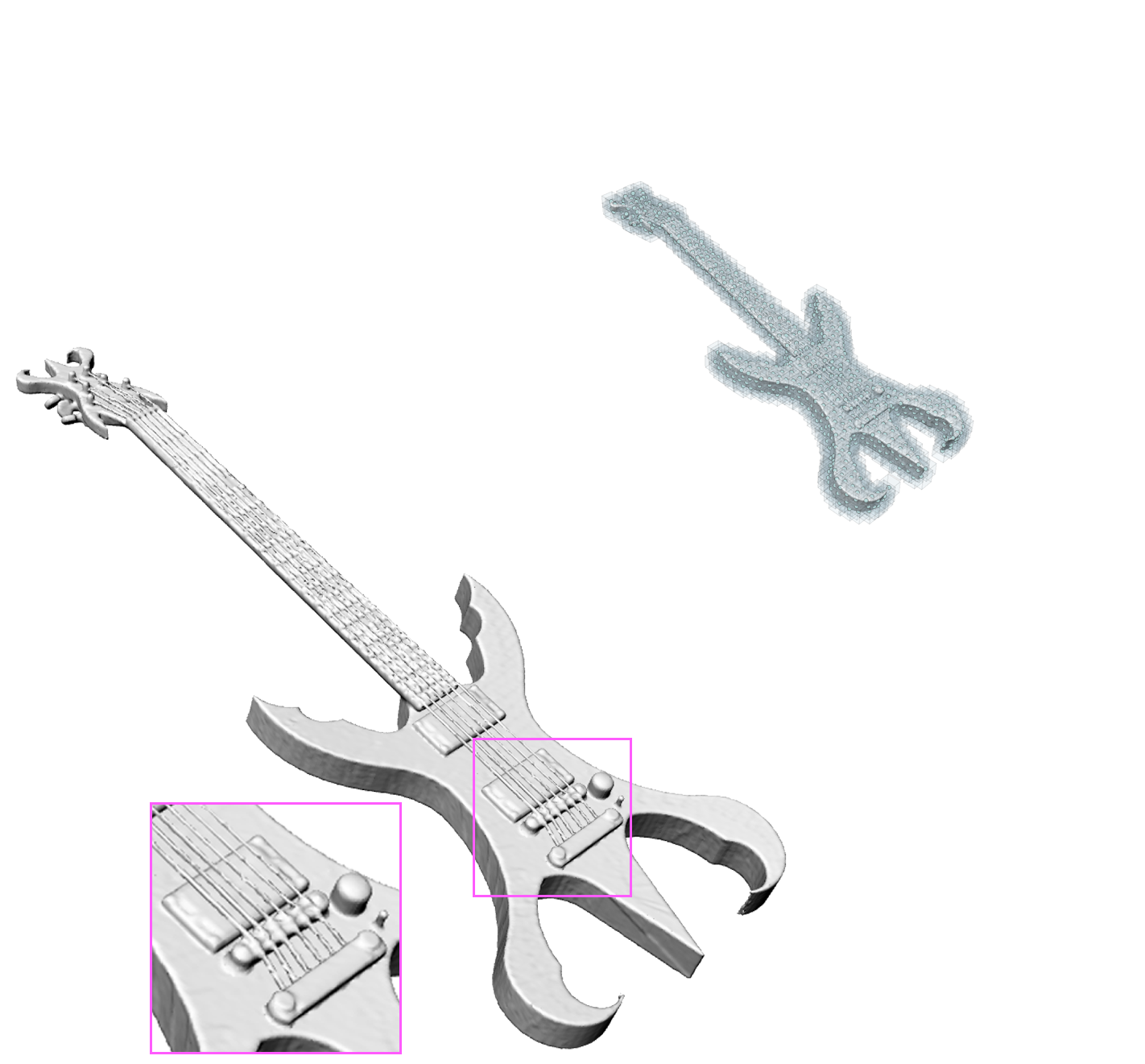} & 
\includegraphics[width=0.2\textwidth]{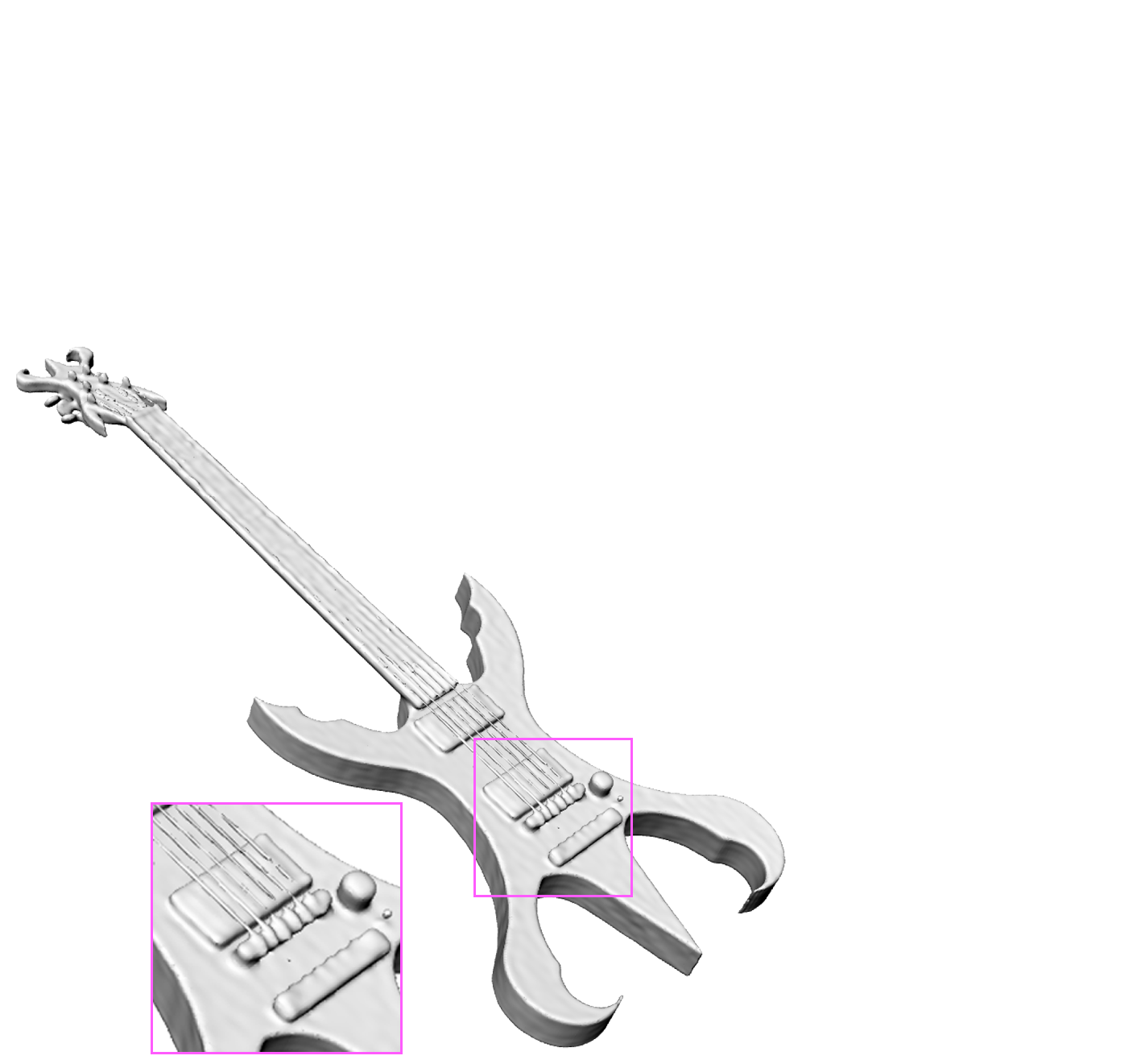} & 
\includegraphics[width=0.2\textwidth]{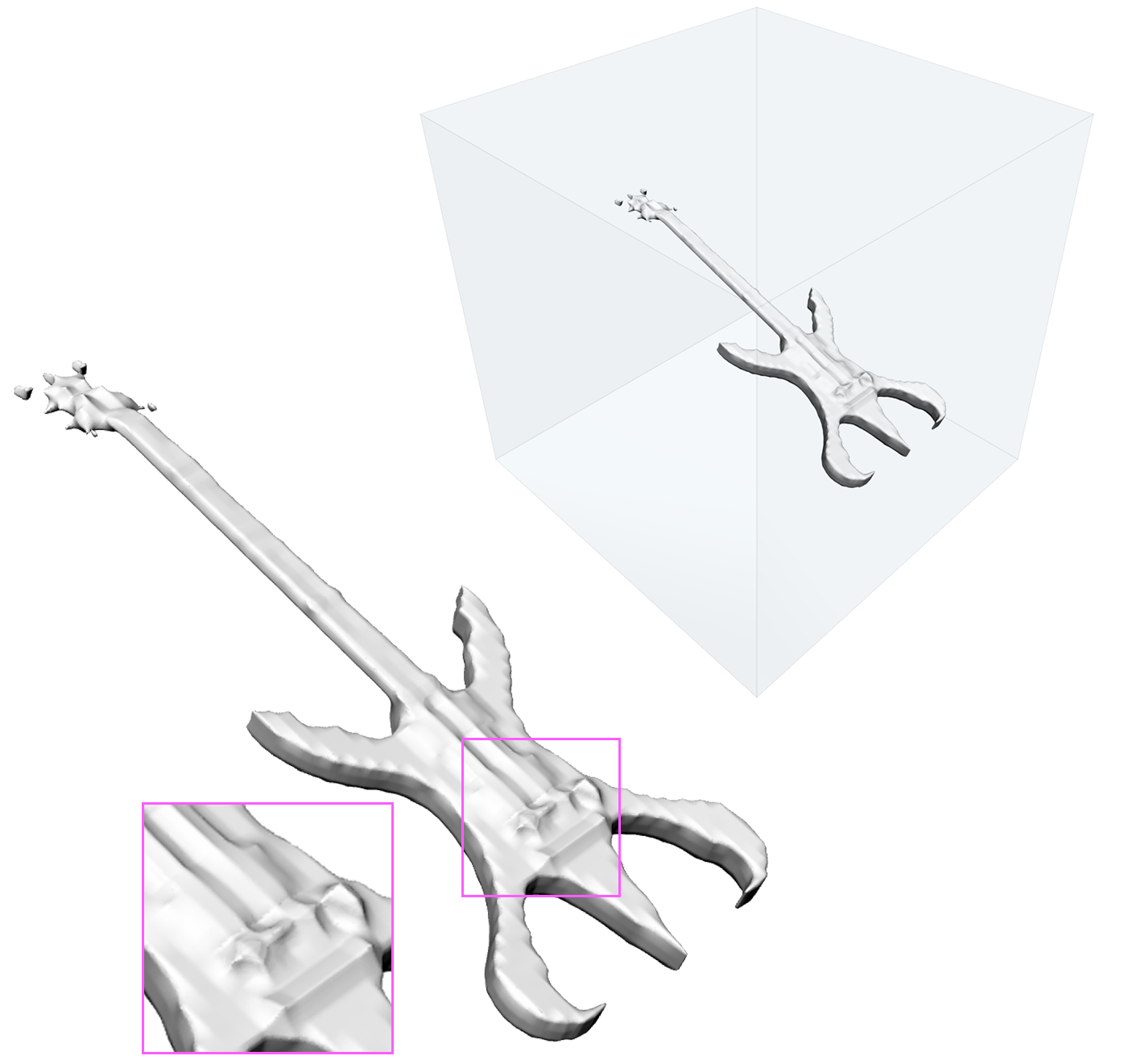} & 
\includegraphics[width=0.2\textwidth]{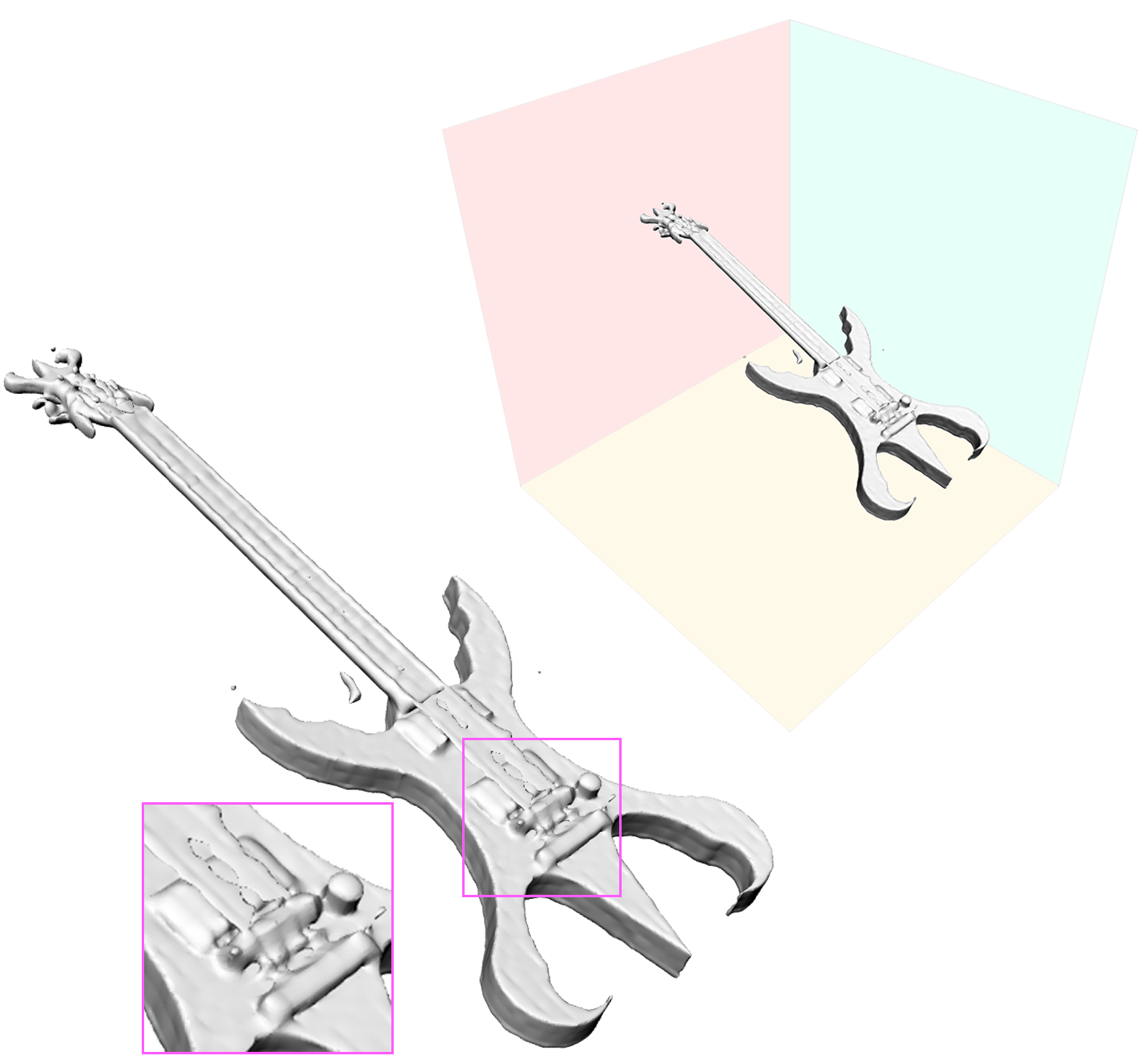} \\
    GT & M-SDF (1 min) & INR (30 min) & 3D Grid (0.5 min) & Triplane (6 min)
\end{tabular}  
\vspace{-5pt}
    \caption{M-SDF representation: we compare the ground truth shape (left), with zero levelsets of (from left to right) M-SDF, Implicit Neural Representation (INR), Volumetric 3D Grid, and Triplane. All representations adhere the same budget of 355K parameters. Note that M-SDF is provides the highest fidelity with an efficient computation time. }
    \vspace{-3pt}
    \label{fig:msdf}
\end{figure*}

\subsection{Computing M-SDF for Shape Dataset}\label{sec:compute_msdf}
Given an input shape $\gS$ (as defined above) from some dataset of shapes, we would like to compute its Mosaic-SDF representation $X$ such that $F_X\approx F_\gS$ in a neighborhood of the shape's boundary surface,  $\partial\gS$. The computation of $X$ consists of two stages: (i) \emph{Initialization}: where we find $X$ such that the domain of definition of $F_X$ covers the surface of the shape, \ie, $\partial\gS \subset \gD(X)$; and (ii) \emph{Fine-tuning}: where we optimize $X$ to improve approximation $F_X\approx F_\gS$. This algorithm can be applied to each shape individually making it simple to parallelize and is computationally efficient compared to alternatives. The algorithm for converting a shape $\gS$ to its M-SDF representation is summarized in Algorithm \ref{alg:mosaicsdf}. Figure \ref{fig:msdf} compares M-SDF representation and some of the popular existing representations for a \emph{fixed budget} of parameters. Note that M-SDF provides the highest quality approximation while is only $\times$2 slower than the fastest method, \ie, the 3D volumetric grid. Later, in Section \ref{s:representation_eval} we provide a more detailed evaluations and comparisons.  

\paragraph{Initialization.} We assume all shapes' boundaries $\partial\gS$ are provided in a way that allows basic operations like sampling and SDF computation, \eg, as triangular meshes. We normalize all shapes so the $\partial \gS$ fit in the cube $[-1,1]^3$. We initialize the volumes centers $\set{\vp_i}_{i \in [n]}$ using farthest point sampling \cite{eldar1997farthest} over the shape's boundary $\partial\gS$. Second, we set $s_i = s$ for all $i\in [n]$, where $s$ is the minimal value that achieves a full coverage of the shape's boundary, \ie, 
\begin{equation}\label{e:coverage}
    s=\min \set{s >0 \ \big\vert \ \partial\gS\subset \gD(X)} ,
\end{equation}
where $\gD(X)$ is defined in \eqref{e:D}. To initialize the local volumes $\mV_i$ we simply store the corresponding values of the SDF $F_\gS$ at the local grid coordinates, \ie, $\mV_i=F_\gS(\vp_i+s_i\sG)$, where $\sG$ defined in \eqref{e:G}.

\paragraph{Fine-Tuning.} Although the initialization already provides a valid approximation to the shape's SDF it can be further improved with quick fine-tuning. We optimize the initialized Mosaic-SDF representation $X$ to reduce the following loss \cite{Park_2019_CVPR,icml2020_2086} striving to regress the SDF's values and first derivatives:
\begin{equation}\label{e:rep_loss}
    \gL(X) = \gL_1(X) + \lambda \gL_2(X), 
\end{equation}
where 
\begin{align}
    \gL_1(X) &= \frac{1}{m}\sum_{j\in[m]}\norm{ F_X(\vx_j) - F_\gS(\vx_j))}_1,\\ 
    \gL_2(X)&=\frac{1}{m} \sum_{j\in[m]} \norm{ \nabla_{\vx}F_X(\vy_j) -  \nabla_{\vx}F_\gS(\vy_j) }_2,
\end{align}
where $\norm{\cdot}_1$, $\norm{\cdot}_2$ represent the 1 and 2-norms (resp.), $\lambda>0$ is a hyper-parameter. The sampling points $\set{\vx_i}_{j\in [m]}$, $\set{\vy_j}_{j\in [m]}$ are spread uniformly over the shapes' boundaries $\partial \gS$ and their neighborhood; more details are in Section \ref{s:implementation_details}.  \vspace{-20pt}



\begin{algorithm}
\DontPrintSemicolon
 \caption{Mosaic-SDF preprocess.}\label{alg:mosaicsdf}
  \SetKwFunction{FarthestSampling}{FarthestSampling}
  \SetKwFunction{FindCoverage}{FindCoverage}
  \SetKwFunction{UnitGrid}{UnitGrid}
  \SetKwInOut{Input}{Input}
  \SetKwInOut{Output}{Output}
  \Input{Shape $\gS$, set size $n$, grid resolution $k$, $\lambda\geq 0$
  }
  
{\color{Green} $\triangleright$  \textit{Initialization}}\;
  
  $\set{\vp_i}_{i\in [n]} \leftarrow $ farthest point sampling of $\partial\gS$ 
  
  $\set{s_i}_{i\in [n]} \leftarrow s$ minimal covering scalar {\color{Green}\algorithmiccomment{eq.~\ref{e:coverage}}}
 
  
  \For{$i\leftarrow 1$ \KwTo $n$}{     
    $\mV_i \leftarrow F_\gS(\vp_i + \sG \cdot s_i)$ {\color{Green}\algorithmiccomment{$\sG$ in eq.~\ref{e:G}}}} 

{\color{Green} $\triangleright$  \textit{Fine-tuning}}\;

  $X \leftarrow  \set{ (\vp_i, s_i, \mV_i)}_{i\in [n]}$\;
  \While{not converged}{
  Take gradient step with $\nabla_X \gL(X)$  
    {\color{Green}\algorithmiccomment{eq.~\ref{e:rep_loss}}}  
  }    
  \Output{$X$}
\end{algorithm} \vspace{-10pt}

\subsection{Mosaic-SDF Generation with Flow Matching}
At this point we assume to be a given a dataset of $N$ shapes paired with condition vectors (\eg, classes or text embeddings), $\set{(\gS^i,\vc^i)}_{i\in [N]}$, and in a preprocess step we converted, using Algorithm \ref{alg:mosaicsdf}, all shapes to M-SDF form, $\set{(X^i,\vc^i)}_{i\in[N]}$. Our goal is to train a flow based generative model taking random noise to M-SDF samples. Given an M-SDF sample $X$ the shape's boundary can be extracted via zero level set contouring of $F_X$, \eg, with Marching Cubes \cite{lorensen1998marching}. Below we recap the fundamentals of flow-based models adapted to our case and present the full training (Algorithm \ref{alg:fm_training}) and sampling (Algorithm \ref{alg:sampling}).

\paragraph{Flow-based generative model.} Flow-based generative models are designed to find a transformation taking samples from some simple noise distributions $X_0\sim p(X_0)$ to conditioned data samples $X_1\sim q(X_1|\vc)$. Our noise distribution is, as customary, a standard Gaussian $p=\gN(0,I)$, and our empirical target distribution is
\begin{equation}
    q(\cdot,\vc) = \frac{1}{N}\sum_{i\in [N]\vert \vc^i=\vc} \delta(\cdot - X^i),
\end{equation}
where $\delta(X)$ is a near-delta function (\eg, a Gaussian with small standard deviation $\sigma$) centered at $0$. In our case, the data points are matrices, 
\begin{equation}\label{e:X}
    X\in \Real^{n\times d}, \text{ where } d=3+1+k^3,
\end{equation}
with their row order considered only up to a permutation. 
A flow model is modeled with a \emph{learnable velocity field} \cite{chen2018neural} denoted $U^\theta:[0,1]\times \Real^{n\times d}\times \Real^c \too \Real^{n\times d}$, where $\theta$ represents its learnable parameters, and $c$ the dimension of an optional conditioning variable. Sampling from a flow model represented by $U^\theta$ is done by first randomizing $X_0\sim p(X_0)$, and second, solving the ODE
\begin{equation}\label{e:ode}
    \dot{X}_t = U^\theta_t(X_t,\vc),   
\end{equation}
with initial condition $X_0$, from time $t=0$ until time $t=1$, and $\vc\in\Real^c$ is the condition vector. Lastly, the desired sample of the model is defined to be $X_1$. 

\paragraph{Symmetric Data.}
As mentioned above $X\in\Real^{n\times d}$ represents a set of $n$ elements in $\Real^d$ (\ie, the elements are the rows of $X$). Differently put, permutation of the rows of $X$, \ie, $\mP X$ with $\mP$ being a permutation matrix, is a \emph{symmetry} of this representation, namely represents the same object $X\cong \mP X$. Consequently, we would like our generative model to generate $X$ and $\mP X$ with the \emph{same probability}. One way to achieve this is to consider noise density $p$ that is \emph{permutation invariant}, \ie,
\begin{equation}\label{e:invariant_p}
    p(\mP X) = p(X), \quad \text{for all } X,\mP,
\end{equation}
and a \emph{permutation equivariant} flow field $U^\theta$, \ie, 
\begin{equation}\label{e:equivariant_U}
    U^\theta_t(\mP X, \vc) = \mP U_t^\theta(X, \vc), \quad \text{for all } t,X,\vc,\mP.
\end{equation}
Indeed as proved in \cite{kohler2020equivariant} (Theorem 1 and 2) equations \ref{e:invariant_p} and \ref{e:equivariant_U} imply that the generations $X(1)$ using an equivariant model $U^\theta$ and invariant noise $p$ are also permutation equivariant and $X(1)$ and $\mP X(1)$ are generated with the same probability, as required. One benefit of this set symmetry is the existence of a powerful permutation equivariant neural architecture, namely a Transformer \cite{vaswani2017attention,nichol2022pointe} without positional encoding.

\paragraph{Flow Matching.}
We use the recent formulation of Flow Matching (FM) \cite{lipman2023flow} with its permutation equivariant variant \cite{klein2023equivariant}. Flow Matching models are similar to diffusion models in taking noise to data but directly regress the velocity field of a noise-to-data target flow and consequently have several benefits such as flexibility of noise-to-data paths, they are well defined for the entire time interval from noise to data (\ie, work with 0 SNR noise), easier to sample due to lower kinetic energy \cite{shaul2023kinetic}, and provide a competitive generation quality. We train Flow Matching with Classifier Free Guidance (CFG) \cite{ho2022classifier} by minimizing the loss
\begin{align}
   \gL(\theta) &= \E_{t,b,p(X_0),q(X_1,\vc)} \norm{U^\theta_t(X_t,\bar{\vc}(b)) - \dot{X}_t}^2
\end{align}
where $t\sim \gU([0,1])$ is the uniform distribution, $b\in\gB(p_{\text{uncond}})$ is a Bernoulli random variable taking values $0,1$ with probability $(1-p_{\text{uncond}}),p_{\text{uncond}}$ (resp.), $\bar{\vc} = (1-b)\cdot \vc + b \cdot \varnothing$ where $\varnothing$ is a symbol of null conditioning, and $X_t$ is a path interpolating noise $X_0$ and data $X_1$. We opt for paths $X_t$ that form Optimal Transport displacement map \cite{mccann1997convexity} conditioned on a training sample $X_1\sim q(X_1)$, \ie, 
\begin{equation}\label{e:Xt}
    X_t = (1-\rho t)X_0 + t X_1,\quad  \rho=1-\sigma
\end{equation}
where $\sigma>0$ is a hyper-parameter chosen to be $\sigma=10^{-5}$ in our case. This path choice is referred to as Conditional Optimal Transport (Cond-OT) and it takes samples from $p(X_0)$ to samples from $\gN(X_1, \sigma^2I)$. Equivalent formulations of Flow Matching are also introduced in \cite{albergo2022building,liu2022flow}.

\begin{algorithm}[t]
\DontPrintSemicolon
\caption{Flow Matching training.}
\label{alg:fm_training}
\SetKwInOut{Input}{Input}
\SetKwInOut{Output}{Output}
\Input{ M-SDF dataset $\set{X^i}_{i\in [N]}$, 
$p_\text{uncond}$, $\sigma$}
Initialize $U_t^\theta$ \; 
\While{not converged}{
$t \sim \gU([0,1]) $ {\color{Green}\algorithmiccomment{sample time}}\;
$(x_1,\vc)\sim q(x_1,\vc)$ {\color{Green}\algorithmiccomment{sample data and condition}}\;
$\vc\leftarrow \varnothing$ with probability $p_\text{uncond}$ {\color{Green}\algorithmiccomment{null condition}} \;
$X_0\sim p(X_0)$ {\color{Green}\algorithmiccomment{sample noise}}\;
$X_t\leftarrow t X_1 + (1-\rho t) X_0$ {\color{Green}\algorithmiccomment{eq.~\ref{e:Xt}}}\; 
$\dot{X}_t\leftarrow X_1 - \rho X_0$ \;
Take gradient step on $\nabla_\theta \|U^\theta_t(X_t,\vc) - \dot{X}_t   \|^2$
}
\Output{$U^\theta$}
\end{algorithm}

\begin{algorithm}[t]
\DontPrintSemicolon
\caption{ODE sampling.}
\label{alg:sampling}
\SetKwInOut{Input}{Input}
\SetKwInOut{Output}{Output}
\Input{trained model $U^\theta$, condition $\vc$,
\qquad guidance parameter $\omega$, number of steps $m$} 
$ X_0 \sim p(X_0)${\color{Green}\algorithmiccomment{sample noise}} \;
$h\leftarrow \frac{1}{m}$ {\color{Green}\algorithmiccomment{step size}}\;
$V_t(\cdot) \leftarrow (1+\omega) U^\theta_t(\cdot|\vc) -  \omega U^\theta_t(\cdot|\varnothing) ${\color{Green}\algorithmiccomment{CFG velocity}}\;
\For{$t = 0, h, \ldots, 1-h$}{    
    $X_{t+h} \leftarrow $ \text{ODEStep}($V_t$, $X_t$) {\color{Green}\algorithmiccomment{ODE solver step}}\;
}
\Output{$X_1$}
\end{algorithm}

%% file: sec/4_experiments.tex
\begin{figure*}
    \centering
    \includegraphics[width=\textwidth]{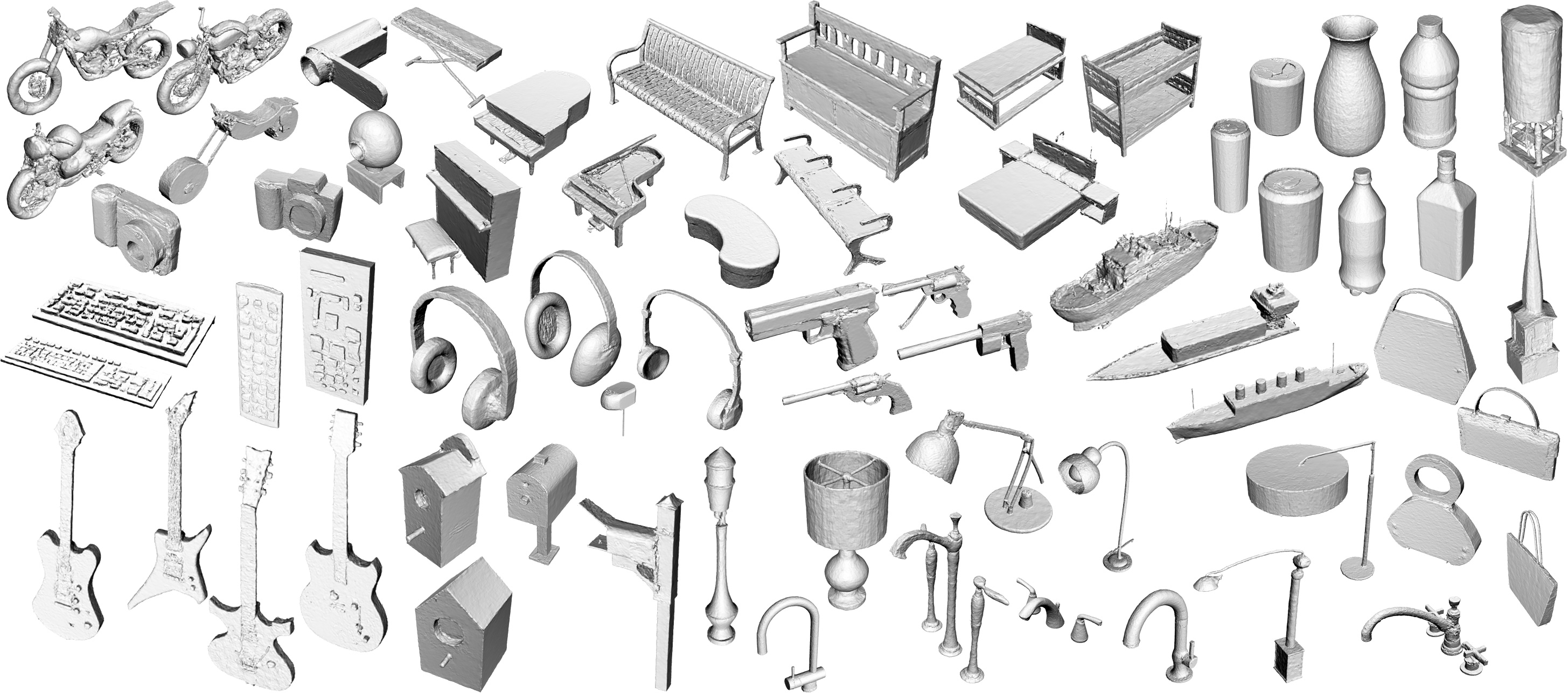}\vspace{-3pt}
    \caption{Conditional samples from a Flow Matching model trained on M-SDF representations of ShapeNetCore v2. \vspace{-7pt}}
    \label{fig:shapenet_generation}
\end{figure*}

\section{Experiments}
\label{sec:experiments}

\subsection{Implementation details}
\label{s:implementation_details}
\paragraph{Datasets and preprocess.}
We train our generative model on two main datasets: 3D Warehouse data, commonly referred to as ShapenetCore v2 \cite{shapenet2015} consisting of 50K 3D polygonal meshes classified to 55 different categories, and a dataset of 600K polygonal meshes with matching text descriptions \cite{luo2023scalable}. We preprocess each polygonal mesh as follows: First, scale it to be bounded inside the unit cube $[-1,1]^3$. Second, make it watertight using \cite{huang2018robust, point-cloud-utils}. Third, we run Algorithm \ref{alg:mosaicsdf} for 1K iterations, with $n=1024$, $k=7$, and $\lambda=0.1$. Consequently the M-SDF tensor representation is $X\in\Real^{1024 \times (3+1+7^3)}$ (see \eqref{e:X}) and has 355K parameters in total. This last step takes less than 2 minutes on a single Nvidia A100 GPU (it takes a bit longer than the experiment in Figure \ref{fig:msdf} since here we used $\lambda>0$). Fourth, noting that the M-SDF representation has three channels, $(\vp_i,s_i,\mV_i)$, we normalize $\vp_i,s_i$ to have zero mean and unit max norm (using 50K random samples of each channel). 

\vspace{-5pt}
\paragraph{Flow Matching model architecture and training.} We train a Flow Matching generative model \cite{lipman2023flow} where for $U^\theta$ we use the transformer-based architecture \cite{vaswani2017attention} without positional encoding to achieve permutation equivariance, compatible with our M-SDF tensorial representation. Each element in the set (\ie, row) of the noisy sample $X_t \in \Real^{n\times d}$ is fed in as token, as well as the time $t$ and the conditioning $\vc$. Our transformer is built with 24 layers with 16 heads and 1024 hidden dimension, which result in a 328M parameter model. We train $U^\theta$ for 500K iterations with batch size of 1024 using the \textsc{Adam} optimizer \cite{kingma2014adam} and learning rate of $1\mathrm{e}{-4}$ with initial warm-up of 5K iterations. We additionally perform EMA (Exponential Moving Average) to the transfomer's weights. Both training were done on 8 nodes of 8 NVIDIA A100 GPUs, which takes around a week.

\subsection{Representation evaluation}
\label{s:representation_eval}
We start with comparing M-SDF to existing popular SDF representations used in 3D generative models focusing on two key aspects: preprocess efficiency, and parameter efficiency. We only consider SDF representations computed independently for each individual shape, \ie, does not use latent space representations defined by a global encoder/decoder. The main reason for this choice is that all methods, including M-SDF, can be adapted to work on latent space, which is an orthogonal design choice. We compare to: 3D Volumetric Grid (3D-Grid), Triplane and Implicit Neural Representation (INR). 

\begin{figure}
\centering
\begin{tabular}{@{\hspace{-10pt}}c@{\hspace{1pt}}c@{\hspace{1pt}}} 
\includegraphics[width=0.5\columnwidth]{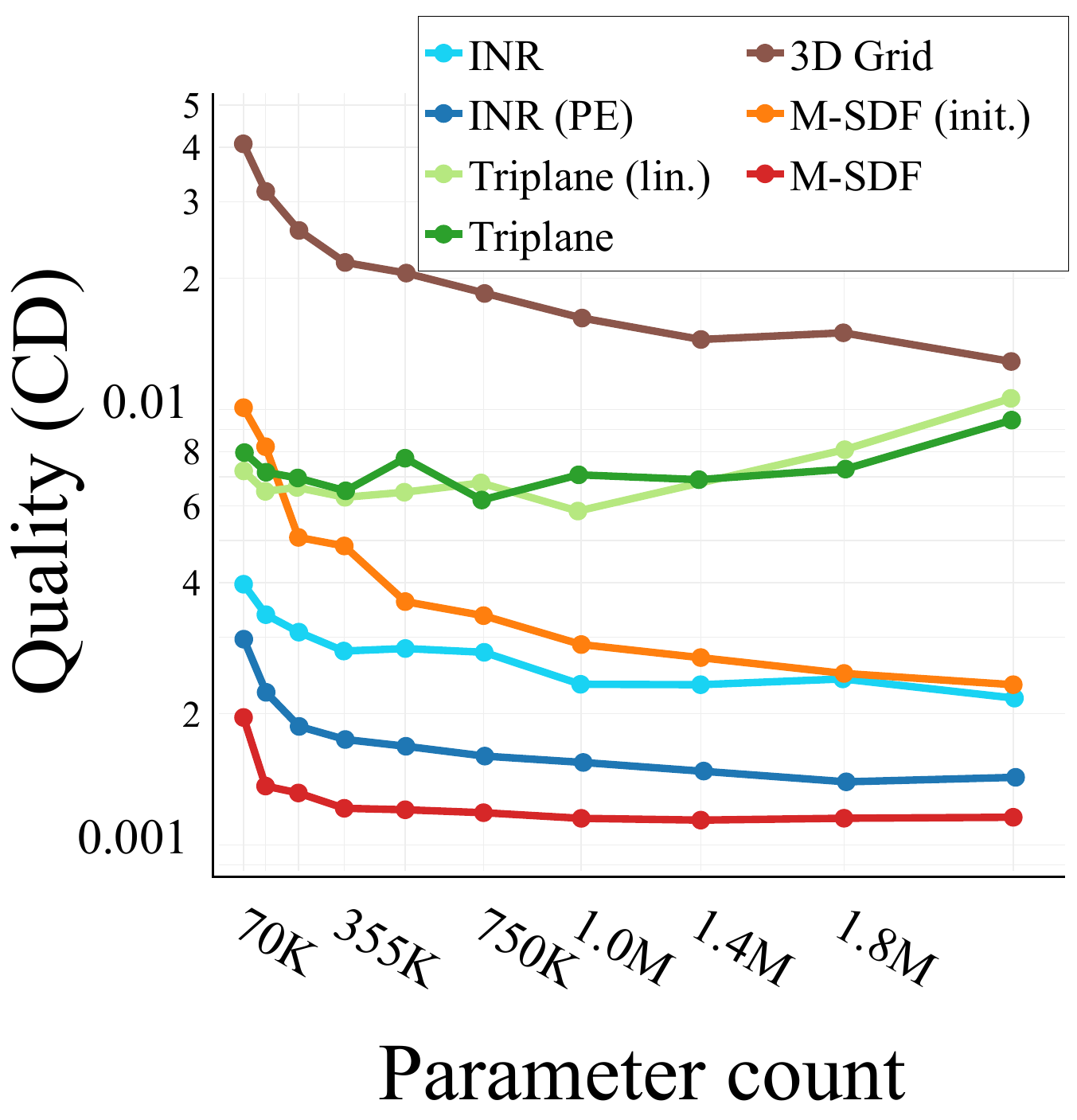} & \includegraphics[width=0.5\columnwidth]{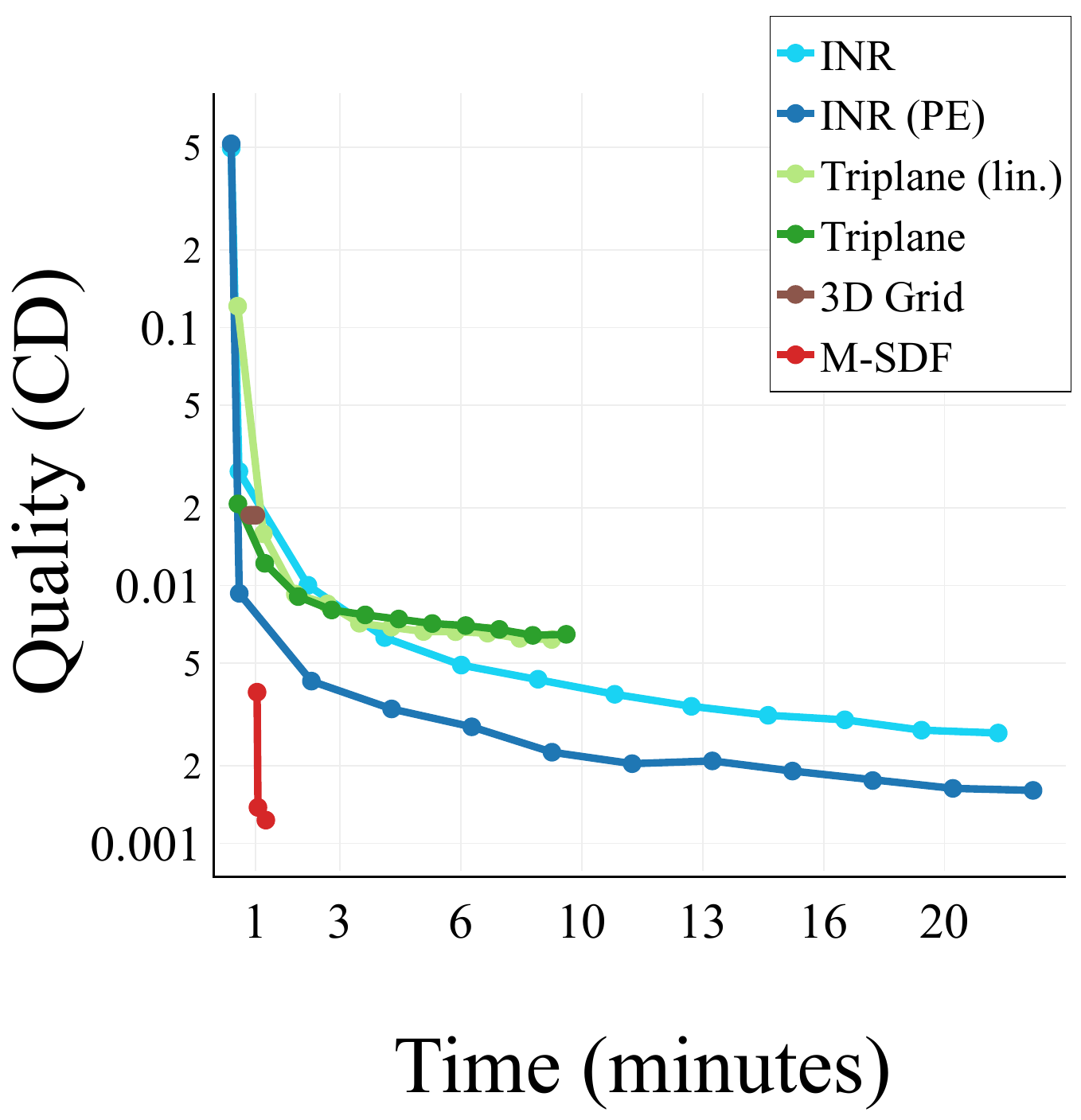} \\
    (a)  & (b)  \vspace{-6pt}
\end{tabular}    
    \caption{(a) 3D approximation quality vs.~representation parameter count; (b) pre-process training time vs.~3D approximation quality for a fixed representation budget of 355K parameters. \vspace{-10pt}  }
    \label{fig:rep_eval}
\end{figure}

We consider 100 random (GT) models from ShapeNetCore-V2 (3D Warehouse) and for each SDF representation (M-SDF, 3D-Grid, Triplane, INR) we log its average preprocess time and surface approximation quality for varying parameter budget. The preprocess time is measured as the wall-clock time it takes a single Nvidia A100 GPU to compute the representation. For Triplane, INR and M-SDF we use the loss in \eqref{e:rep_loss} with $\lambda=0$; 3D-Grid is computed by evaluating the GT SDF at the grid nodes. The surface approximation quality is measured by the Chamfer Distance (CD) between the extracted surface mesh from the SDF representation and the GT mesh. 
Figure \ref{fig:rep_eval} summarizes the results in two graphs: (a) shows the surface approximation quality of each representation for different parameter budgets; and (b) surface approximation quality versus preprocess time of each representation.
For INR we additionally examine the option of using Positional Encoding (PE) \cite{mildenhall2020nerf}, which incorporates high frequencies as input to the MLP. For Triplane we evaluated the two alternatives of aggregating the projected features, as suggested in \cite{kplanes_2023}, using learned linear decoder (denoted as lin.) or using a small MLP decoder. We note that while previous works that use the Triplane representation \cite{shue20223d, Chan2021} suggested additional regularizations, we tested Triplane with the same supervision and loss as the other methods that require optimization, namely INR and M-SDF. This can potentially explain the degradation in the approximation quality as the parameter count increases. Additionally, for M-SDF we report the surface approximation quality both at initialization and after fine-tuning (see Section \ref{sec:compute_msdf}).
As can be seen in the graphs, M-SDF is superior to INR in terms of surface approximation per parameter budget while is computable in only a fraction of the time. 3D Grids are the only faster baseline to M-SDF but their approximation-parameter trade-off is considerably worse (see also Figure \ref{fig:msdf}).

\input{tables/shapenet_eval_new}

\begin{figure}
\centering
\begin{tabular}{@{\hskip0pt}c@{\hskip3pt}cccc@{\hskip0pt}c@{\hskip0pt}c@{\hskip0pt}}
\rotatebox{90}{\scriptsize \textbf{\textit{3DILG}}} &
\includegraphics[width=0.06\columnwidth]{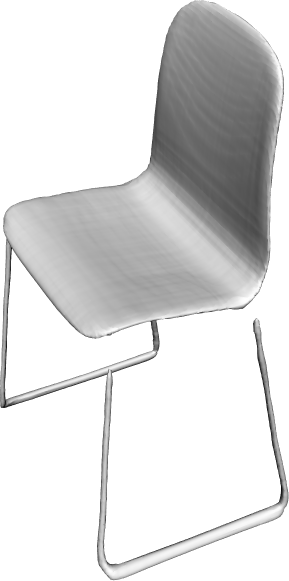} &  
\includegraphics[width=0.10\columnwidth]{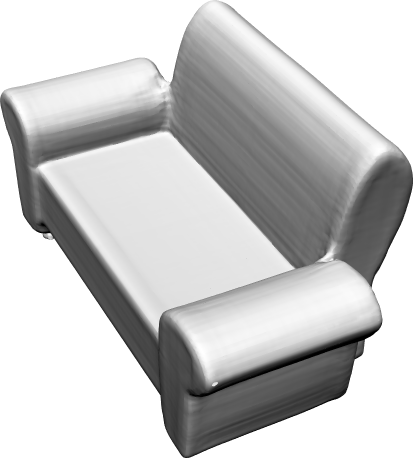} &  
\includegraphics[width=0.10\columnwidth]{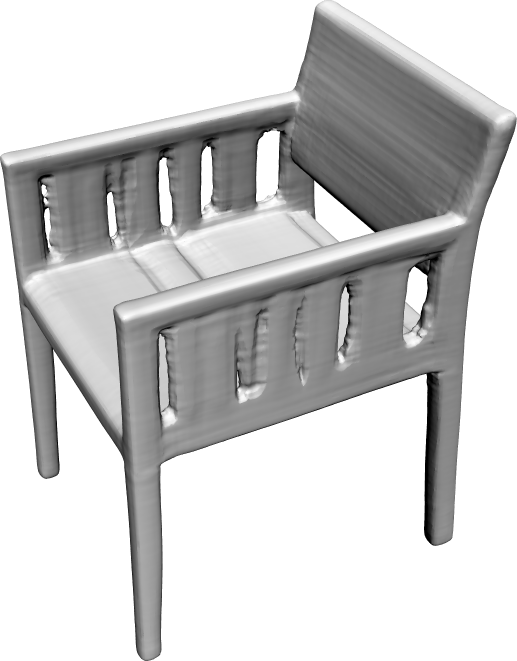} &  
\includegraphics[width=0.18\columnwidth]{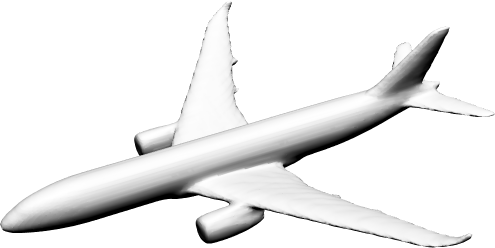} &  
\includegraphics[width=0.18\columnwidth]{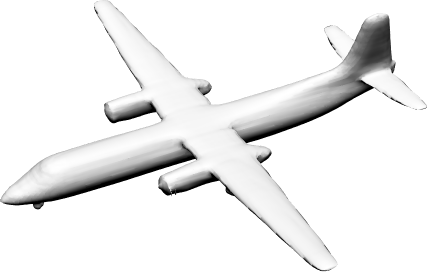} &  
\includegraphics[width=0.18\columnwidth]{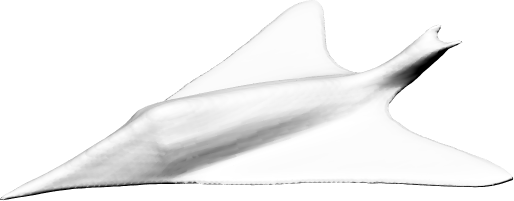} \\ 
\rotatebox{90}{\scriptsize \textbf{\quad \textit{NW}}} &
\includegraphics[width=0.10\columnwidth]{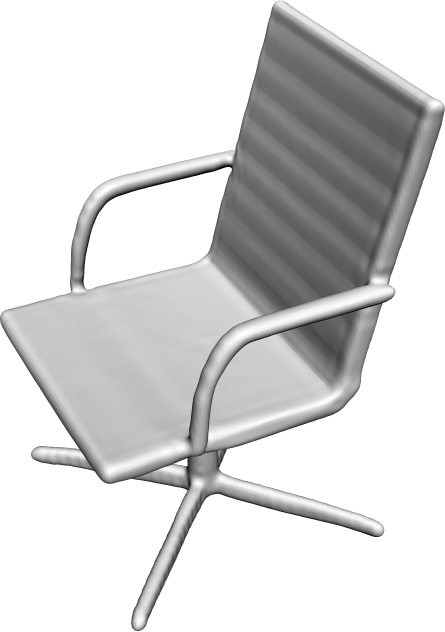} &  
\includegraphics[width=0.10\columnwidth]{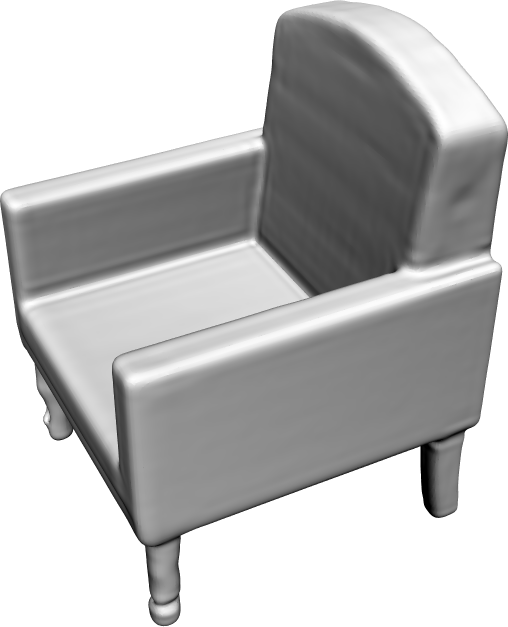} &  
\includegraphics[width=0.07\columnwidth]{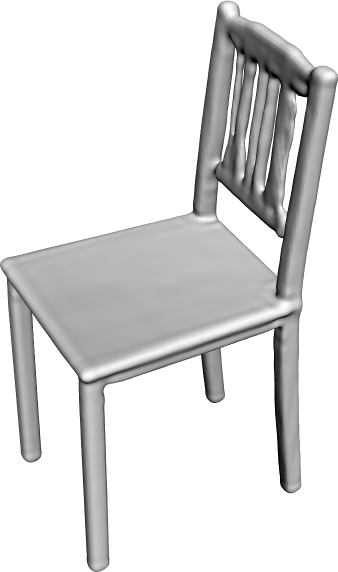} &  
\includegraphics[width=0.18\columnwidth]{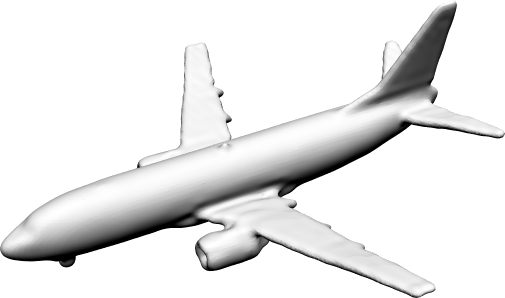} &  
\includegraphics[width=0.18\columnwidth]{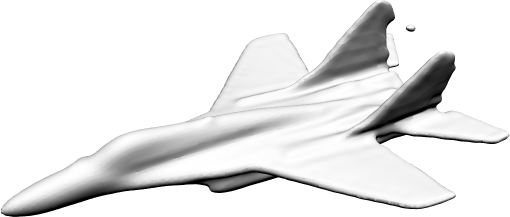} &  
\includegraphics[width=0.18\columnwidth]{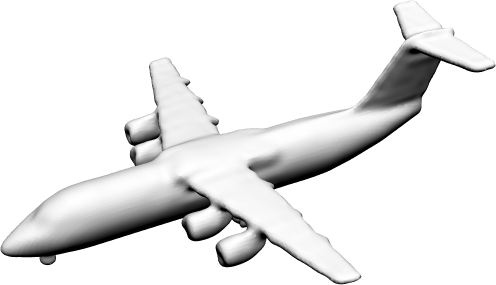} \\  
\rotatebox{90}{\scriptsize \textbf{\quad \textit{S2VS}}} &
\includegraphics[width=0.10\columnwidth]{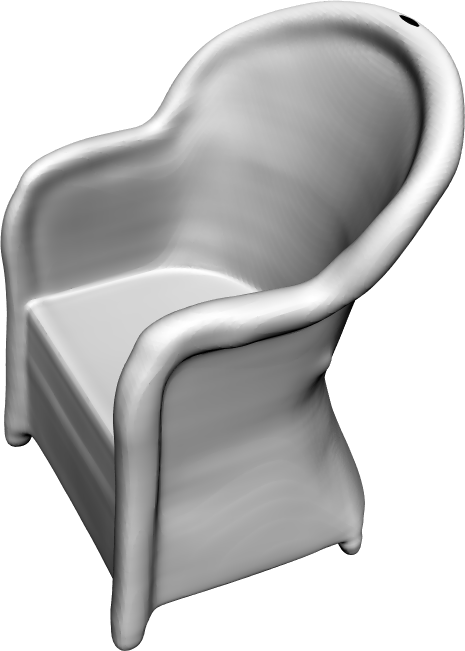} &  
\includegraphics[width=0.10\columnwidth]{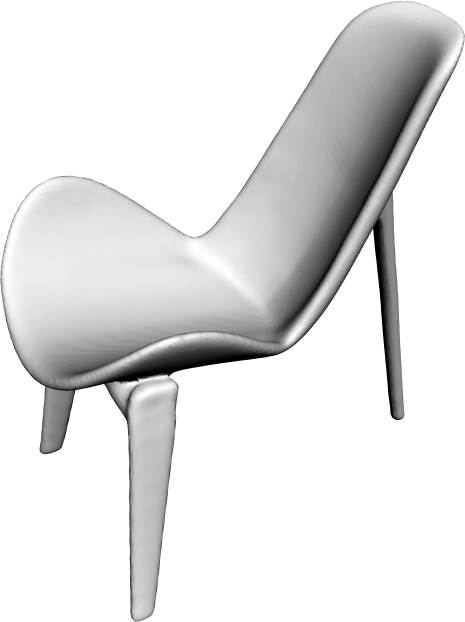} &  
\includegraphics[width=0.10\columnwidth]{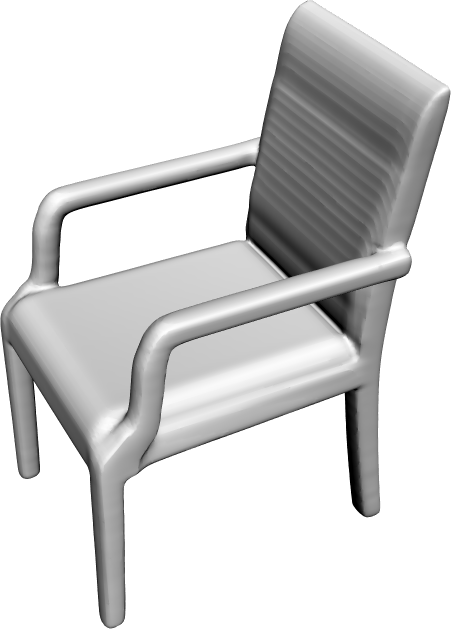} &  
\includegraphics[width=0.18\columnwidth]{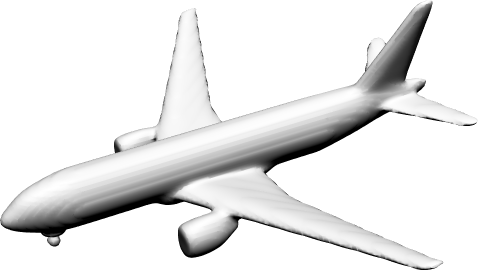} &  
\includegraphics[width=0.18\columnwidth]{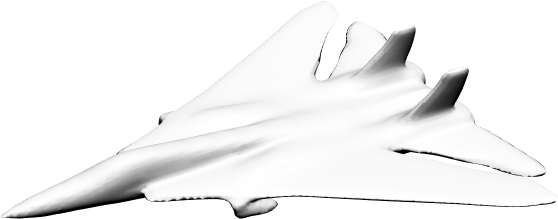} &  
\includegraphics[width=0.18\columnwidth]{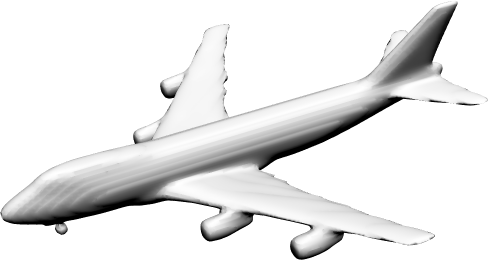} \\ 
\rotatebox{90}{\scriptsize \textbf{\textit{\quad Ours}}} &
\includegraphics[width=0.10\columnwidth]{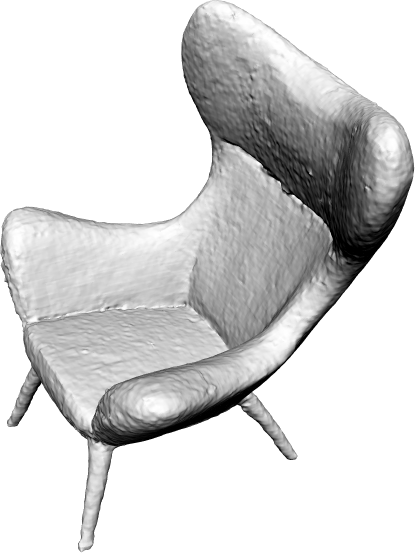} &  
\includegraphics[width=0.10\columnwidth]{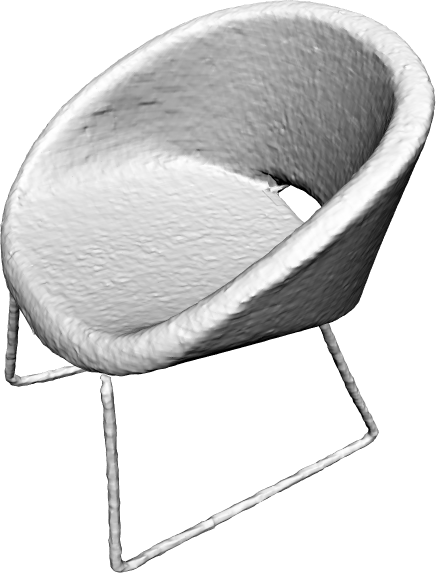} &  
\includegraphics[width=0.10\columnwidth]{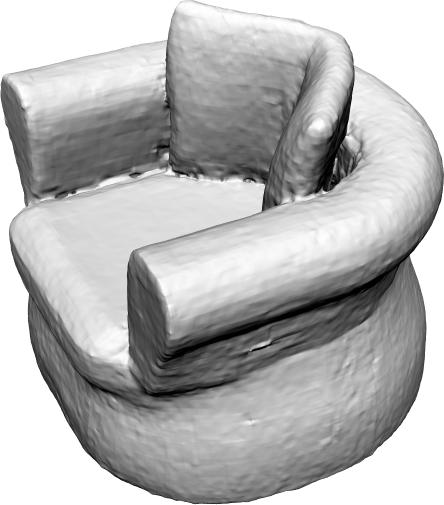} &  
\includegraphics[width=0.18\columnwidth]{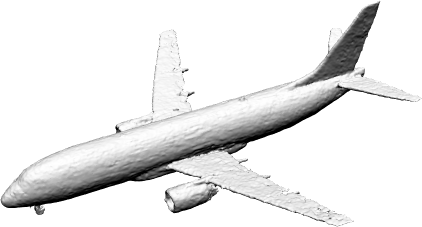} &  
\includegraphics[width=0.18\columnwidth]{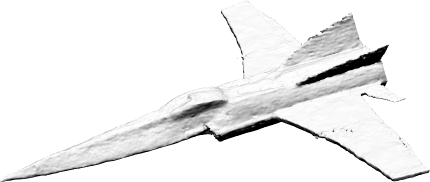} &  
\includegraphics[width=0.18\columnwidth]{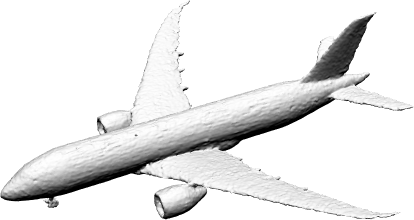}  
\end{tabular}
\vspace{-5pt}
     \caption{Class conditioning generation of 3D shapes compared to relevant baselines. Note the high fidelity demonstrated in M-SDF generated shapes compared to the (overly smooth) baselines. \vspace{-10pt}}\label{fig:shapenet_compare}
\end{figure}

\begin{figure*}
    \centering
    \includegraphics[width=0.98\textwidth]{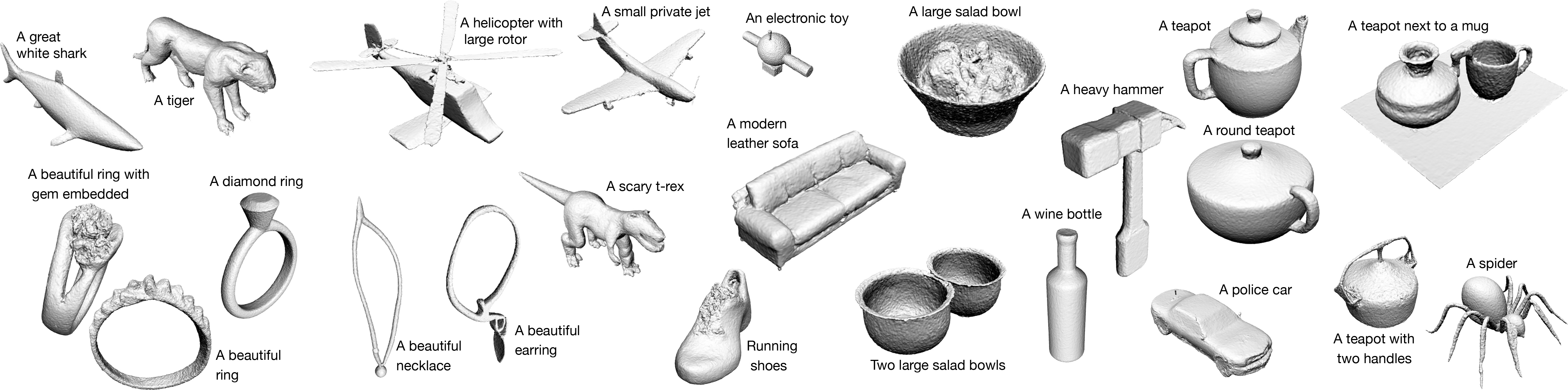}
    \vspace{-3pt}
    \caption{Text-to-3D samples from a Flow Matching model trained on M-SDF representations. \vspace{-7pt}}
    \label{fig:text_to_3d_generation}
\end{figure*}

\subsection{Class conditional generation}\label{sec:class_gen}
In this section we evaluate our class-conditioned generative FM model trained on the ShapeNetCore-V2 (3D Warehouse) \cite{shapenet2015} where the 55 classes of objects in this dataset are used as conditions. We follow the split to train/val/test suggested by \cite{zhang2022dilg}. Following and extending protocols of past works, Table \ref{tab:shapenet_cond} reports quantitative evaluation for the 5 largest classes (containing over 3K shapes each) comparing M-SDF and relevant recent baselines.

\paragraph{Evaluation metrics.} We make use of several standard metrics for evaluating the performance of our M-SDF generative models. All these metrics quantify  differences between a set of reference shapes $S_r$ and generated shapes $S_g$.
To measure distances between shapes we follow previous works (\eg, \cite{zeng2022lion,luo2021diffusion,zhang20233dshape2vecset}) and use the Chamfer Distance (CD) and Earth Moving Distance (EMD). Using these distances we compute: Maximum Mean Discrepancy (MMD), Coverage (COV), and  1-nearest neighbor accuracy (1-NNA) to quantify fidelity, diversity and distributional similarity, respectively. 
%
%
Furthermore, following \cite{zhang20233dshape2vecset, nichol2022pointe} we use the 3D analogs of the Fréchet Inception Distance (FID) and Kernel Inception Distance (KID), commonly used to evaluate image generative models. We employ a pre-trained PointNet++ \cite{qi2017pointnetplusplus} to extract features.
As in \cite{zhang20233dshape2vecset} we will refer to these metrics as Fréchet PointNet++ Distance (FPD) and Kernel PointNet++ Distance (KPD).
Additional information and implementation details are provided in the supplementary material. 

\paragraph{Baselines.} We compare to Neural Wavelet (NW) \cite{hui2022wavelet} that generate a grid-based representation. Note that this works trains an unconditional model for each class separately, making it arguably simpler than a single conditioned model on the entire 55 classes. We additionally compare to 3DILG \cite{zhang2022dilg} and 3DShape2VecSet (S2VS) \cite{zhang20233dshape2vecset}, that suggest to generate structured or unstructured latent vectors, encoding an occupancy field using a transformer.

As can be read in Table \ref{tab:shapenet_cond} our M-SDF based generative model compares favorably to the baselines, achieving best, or second best performance in all metrics. Figure \ref{fig:shapenet_compare} shows qualitative comparison for two classes common for all baselines, \ie, chairs and airplanes. Note that generation with M-SDF allows higher fidelity and  sharper surfaces compared to baselines which tend to be overly smooth. Figure \ref{fig:shapenet_generation} shows additional class conditional generations with our M-SDF model. 

\input{tables/mc_time}

\paragraph{SDF evaluation time.} 
An additional advantage of M-SDF compared to other performant baselines is the relative efficiency and flexibility in extracting the zero level set of the SDF $F_X$ (\eqref{e:F_X}). 3DILG and S2VS require forward pass in a transformer for function evaluation. NW is restricted to generate a 3D grid in a single resolution. M-SDF can be evaluated efficiently by querying only the relevant local grids. In Table \ref{tab:mc_time} we log, for each method, the total time of extracting the zero levelset of the SDF (with Marching Cubes) using cubic grids of different resolutions ($64^3$, $128^3$, $256^3$, and $512^3$); OOM stands for Out of Memory.

%

\subsection{Text-to-3D generation}\label{s:textcond}
Lastly we provide a qualitative evaluation of our M-SDF based generative FM model trained on a dataset of 600K shapes with matching text captions \cite{luo2023scalable}. We utilize a pre-trained text model \cite{2020t5} as our textual embedding, passing this embedding to the model as conditioning. Figure \ref{fig:text_to_3d_generation} depicts pairs of generated shapes and the text conditions used to generate it.


%% file: tables/shapenet_eval_new.tex
\renewcommand{\arraystretch}{1.0} 
\begin{table}[h]
    \centering
    \resizebox{1.\linewidth}{!}{
    \begin{tabular}{lcccccccc} 
     \toprule
       &  FPD ($\downarrow$) & KPD ($\downarrow$) &  \multicolumn{2}{c}{COV ($\uparrow$,\%)} & \multicolumn{2}{c}{MMD ($\downarrow$)} & \multicolumn{2}{c}{1-NNA ($\downarrow$,\%)}\\
       &   &  &  \small{CD}  & \small{EMD} & \small{CD} & \small{EMD} & \small{CD} & \small{EMD} \\
       \bottomrule
 airplane & & & & & & & &      \\ \midrule
\textbf{\textit{3DILG}} & 1.83 & 3.22 & 41.09 & 32.67 & 4.69 & 4.73 & 82.67 & 84.41 \\
\textbf{\textit{NW}} & \cellcolor{lightgray} {0.81} & \cellcolor{lightgray} {1.26} & \cellcolor{lightgray} {51.98} & \cellcolor{lightgray} {45.05} & \cellcolor{lightblue} {3.36} & \cellcolor{lightgray} {4.19} & \cellcolor{lightgray} {68.32} & \cellcolor{lightgray} {73.76} \\
\textbf{\textit{S2VS}} & 0.94 & 1.65 & 51.98 & 40.59 & 3.80 & 4.45 & 69.06 & 76.73 \\
\textbf{\textit{Ours}} & \cellcolor{lightblue} {0.44} & \cellcolor{lightblue} {0.50} & \cellcolor{lightblue} {52.48} & \cellcolor{lightblue} {51.49} & \cellcolor{lightgray} {3.54} & \cellcolor{lightblue} {3.78} & \cellcolor{lightblue} {62.62} & \cellcolor{lightblue} {69.55} \\
       \bottomrule
 car & & & & & & & &      \\ \midrule
\textbf{\textit{3DILG}} & 2.84 & 6.24 & 18.86 & 20.57 & 4.67 & 3.83 & 93.43 & 90.57 \\
\textbf{\textit{NW}} & -  & -  & -  & -  & -  & -  & -  & -  \\
\textbf{\textit{S2VS}} & \cellcolor{lightgray} {1.32} & \cellcolor{lightgray} {2.17} & \cellcolor{lightgray} {37.71} & \cellcolor{lightgray} {40.00} & \cellcolor{lightgray} {4.13} & \cellcolor{lightgray} {3.52} & \cellcolor{lightgray} {84.57} & \cellcolor{lightgray} {86.57} \\
\textbf{\textit{Ours}} & \cellcolor{lightblue} {0.46} & \cellcolor{lightblue} {0.48} & \cellcolor{lightblue} {45.71} & \cellcolor{lightblue} {51.43} & \cellcolor{lightblue} {2.87} & \cellcolor{lightblue} {2.75} & \cellcolor{lightblue} {70.00} & \cellcolor{lightblue} {66.00} \\
       \bottomrule
 chair & & & & & & & &      \\ \midrule
\textbf{\textit{3DILG}} & 1.64 & 2.00 & 37.87 & 39.94 & 20.37 & 10.54 & 74.11 & 69.38 \\
\textbf{\textit{NW}} & 1.41 & 1.29 & 43.79 & 47.04 & \cellcolor{lightgray} {16.53} & 9.47 & 59.47 & 64.20 \\
\textbf{\textit{S2VS}} & \cellcolor{lightgray} {0.77} & \cellcolor{lightgray} {0.63} & \cellcolor{lightblue} {51.78} & \cellcolor{lightgray} {52.37} & 16.97 & \cellcolor{lightgray} {9.44} & \cellcolor{lightgray} {58.43} & \cellcolor{lightgray} {60.80} \\
\textbf{\textit{Ours}} & \cellcolor{lightblue} {0.52} & \cellcolor{lightblue} {0.19} & \cellcolor{lightgray} {48.22} & \cellcolor{lightblue} {55.03} & \cellcolor{lightblue} {15.47} & \cellcolor{lightblue} {9.13} & \cellcolor{lightblue} {51.04} & \cellcolor{lightblue} {55.62} \\
       \bottomrule
 sofa & & & & & & & &      \\ \midrule
\textbf{\textit{3DILG}} & 3.19 & 5.83 & 25.95 & 29.11 & 26.41 & 10.71 & 84.81 & 77.85 \\
\textbf{\textit{NW}} & -  & -  & -  & -  & -  & -  & -  & -  \\
\textbf{\textit{S2VS}} & \cellcolor{lightgray} {1.17} & \cellcolor{lightgray} {1.70} & \cellcolor{lightblue} {48.73} & \cellcolor{lightblue} {51.90} & \cellcolor{lightblue} {10.83} & \cellcolor{lightblue} {7.25} & \cellcolor{lightgray} {62.66} & \cellcolor{lightgray} {57.91} \\
\textbf{\textit{Ours}} & \cellcolor{lightblue} {0.63} & \cellcolor{lightblue} {0.62} & \cellcolor{lightgray} {46.20} & \cellcolor{lightgray} {48.10} & \cellcolor{lightgray} {12.43} & \cellcolor{lightgray} {7.60} & \cellcolor{lightblue} {61.71} & \cellcolor{lightblue} {55.70} \\
       \bottomrule
 table & & & & & & & &      \\ \midrule
\textbf{\textit{3DILG}} & 2.86 & 4.13 & 29.45 & 30.88 & 22.96 & 10.18 & 78.27 & 78.74 \\
\textbf{\textit{NW}} & 1.49 & 2.20 & 51.07 & 47.98 & \cellcolor{lightblue} {13.27} & \cellcolor{lightblue} {7.72} & \cellcolor{lightgray} {56.41} & \cellcolor{lightgray} {58.67} \\
\textbf{\textit{S2VS}} & \cellcolor{lightgray} {0.83} & \cellcolor{lightgray} {0.92} & \cellcolor{lightblue} {53.44} & \cellcolor{lightgray} {49.41} & 14.06 & 8.01 & 59.74 & 61.05 \\
\textbf{\textit{Ours}} & \cellcolor{lightblue} {0.47} & \cellcolor{lightblue} {0.21} & \cellcolor{lightgray} {52.97} & \cellcolor{lightblue} {53.21} & \cellcolor{lightgray} {13.49} & \cellcolor{lightgray} {7.74} & \cellcolor{lightblue} {51.31} & \cellcolor{lightblue} {50.59} \\
       \bottomrule
    \end{tabular}
    }\vspace{-5pt}
        \caption{Evaluation of our class conditioning generation model trained on
        3D Warehouse \cite{warehouse} 
        compared to baselines. We report results on the 5 largest classes in the dataset. KPD is multiplied by $10^3$, MMD-CD by $10^3$ and MMD-EMD by $10^2$. \vspace{-0pt}
}\label{tab:shapenet_cond}
\end{table}

%% file: tables/mc_time.tex
\begin{wraptable}[7]{r}{0.5\linewidth}
    \centering
    \vspace{-13pt}
    \resizebox{\linewidth}{!}{
\begin{tabular}{lcccc} 
     \toprule
       & $64^3$ & $128^3$ &  $256^3$ & $512^3$
      \\ \midrule 
     \textbf{\textit{3DILG}} \ 
    &
    0.3
     &
    2.33
    &
    18.44
     &
    159.56
    \\
      \textbf{\textit{NW}} \ 
     &
    -
    &
    -
    &
    0.61
     &
    -
    \\
     \textbf{\textit{S2VS}} \ 
     &
    0.06
    &
    0.36
    &
    OOM
     &
    OOM
    \\
      \textbf{\textit{Ours}} \ 
     &
    0.05
    &
    0.34
    &
    2.74
    &
    21.48
    \\
     \bottomrule 
    \end{tabular} 
    }
    \caption{Surface extraction time (in seconds).
    }\label{tab:mc_time}
\end{wraptable}

%% file: sec/X_suppl.tex
\clearpage
\setcounter{page}{1}
\maketitlesupplementary
\appendix


\section{Generative model evaluation}\vspace{-3pt}
In this section we provide additional information on the experiments described in Section \ref{sec:class_gen}.

\subsection{Metrics}\vspace{-3pt}
We measure distances between shape distributions following previous works \cite{pointflow,zeng2022lion,luo2021diffusion,zhang20233dshape2vecset}. We quantify differences between a set of reference shapes $S_r$ and a set of generated shapes $S_g$. We describe a shape $\gY \in S_r$ as a point cloud of size $N$ sampled from a reference mesh using the farthest point sampling \cite{eldar1997farthest}. Similarly, $\gX \in S_g$ is a point cloud sampled from a generated surface mesh, extracted as the $0$-level set of the SDF (or $0.5$ level set for occupancy function) using the Marching Cubes algorithm \cite{lorensen1998marching}. 

\vspace{-8pt}
\paragraph{Geometric shape similarity.}
The Chamfer Distance (CD) and the Earth Mover Distance (EMD) measure  similarity between two point clouds:
\begin{equation}
    \text{CD}(\gX, \gY) = \sum_{\vx \in \gX} \min_{\vy \in \gY} \lVert x - y \rVert_2^2 + \sum_{\vy \in \gY} \min_{\vx \in \gX} \lVert x - y \rVert_2^2
\end{equation}
\begin{equation}
    \text{EMD}(\gX, \gY) = \min_{\gamma: \gX \rightarrow \gY} \sum_{\vx \in \gX} \lVert x - \gamma(y) \rVert_2
\end{equation}
where $\gamma$ is the bijection between the point clouds, and $N=5K$. In the following, we denote by $D(\gX, \gY)$  distance measure between two point clouds, referring to either CD or EMD. 
\vspace{-8pt}
\paragraph{Geometrical distances between sets of shapes.} The CD and EMD distances between point clouds are used to define the following distances between \emph{sets} of shapes $S_r$ and $S_g$: Coverage (COV) quantifying the diversity of $S_g$ by counting the number of reference shapes that are matched to at least one generated shape; Minimum Matching Distance (MMD) measuring the fidelity of the generated shapes to the reference set; and 1-Nearest Neighbor Accuracy (1-NNA) describing the distributional similarity between the generated shapes and the reference set, quantifying both quality and diversity. Next we provide the mathematical definitions of these distance measures:
\begin{equation}
    \hspace{-2pt}\text{COV}(S_g, S_r)\hspace{-2pt} =\hspace{-2pt} \frac{\lvert \set{\argmin_{\gY \in S_r} D(\gX,\gY) | \gX \in S_g } \rvert}{\lvert S_r \rvert}
\end{equation}
\begin{equation}
    \text{MMD}(S_g, S_r) = \frac{1}{\lvert S_r \rvert} \sum_{\gY \in S_r} \min_{ \gX \in S_g} D(\gX,\gY)
\end{equation}
\begin{equation}
    \hspace{-8pt}\text{1-NNA}(S_g, S_r)\hspace{-2pt} =\hspace{-2pt} \frac{\hspace{-2pt}\sum\limits_{\gX \in S_g}\hspace{-2pt} \mathbb{I} \left[ N_{\gX} \in S_g \right]\hspace{-2pt} +\hspace{-4pt} \sum\limits_{\gY \in S_r}\hspace{-2pt} \mathbb{I} \left[ N_{\gY} \in S_r \right]}{\lvert S_r \rvert + \lvert S_g \rvert}
\end{equation}
where $\mathbb{I}(\cdot)$ is the indicator function and $ N_{\gX}$ is the nearest neighbor of $\gX$ in the set $S_r \cup S_g - \{\gX \} $. 

\vspace{-8pt}
\paragraph{Perceptual distances between sets of shapes.} Alongside the geometric distance-based metrics, we adopt the 3D analogs of the Fréchet Inception Distance (FID) and Kernel Inception Distance (KID) suggested in previous works \cite{zhang20233dshape2vecset, nichol2022pointe}. In the 3D case, FID/KID are computed on the feature sets computed by pushing $N=2046$ point samples into a pre-trained PointNet++ network \cite{qi2017pointnetplusplus}. 
We denote by $\gR$ and $\gG$ the sets of the extracted features from the reference shapes $S_r$ and the generated shaped $S_g$, respectively. We can further define $(\mu_r, \Sigma_r)$ as the the mean and covariance statistics computed from the feature set $\gR$, and similarly $(\mu_g, \Sigma_g)$ for $\gG$.
As in \cite{zhang20233dshape2vecset}, the Fréchet PointNet++ Distance (FPD) and Kernel PointNet++ Distance (KPD) are defined by
\vspace{-3pt}
\begin{equation}
  \hspace{-4pt}  \text{FPD}(S_g, S_r)\hspace{-2pt} = \hspace{-2pt}\lVert \mu_g \hspace{-1pt}-\hspace{-1pt} \mu_r \rVert \hspace{-2pt} + \hspace{-2pt} {\displaystyle{\mathrm{Tr}}}\left(\Sigma_g+\Sigma_r\hspace{-2pt}-\hspace{-2pt}2(\Sigma_g\Sigma_r)^{\scriptscriptstyle{\frac{1}{2}}}\right) 
\end{equation}
\vspace{-12pt}
\begin{equation}
    \text{FPD}(S_g, S_r) = \left( \frac{1}{\lvert \gR \rvert} \sum_{\rvx \in \gR} \max_{\rvy \in \gG} K(\rvx, \rvy ) \right)^2
\end{equation}
where $K(\cdot, \cdot)$ is a polynomial kernel function distance.

\subsection{Computation of distance metrics}\vspace{-3pt}
Following previous works  we compute the geometric distances, \ie, COV, MMD and 1-NNA, with the reference set of shapes, $S_r$, chosen to be the test split; and we generate an equal number of shape from our generated set $S_g$.
The per-class test split given by \cite{zhang2022dilg} consists of $5\%$ of the shapes from each class, with varying numbers of shapes in each class.
We used the following released codes to compute CD\footnote{https://github.com/ThibaultGROUEIX/ChamferDistancePytorch} and EMD\footnote{https://github.com/daerduoCarey/PyTorchEMD}, and computed the distance metrics from the official code of \cite{zeng2022lion}\footnote{https://github.com/nv-tlabs/LION/tree/main}.

For computing the perceptual distances, FPD and KPD, we follow \cite{zhang20233dshape2vecset} and use 1K generated shapes $S_g$, and as the reference set $S_r$ we take 1K randomly sampled shapes from the train split. We utilize a pre-trained PointNet++ model from \cite{Pointnet} to extract the features $\gR$ and $\gG$. 

To run the baselines, we use the official implementation of each method together with the pre-trained model they supply: The per-class unconditional models from Neural Wavelet\footnote{https://github.com/edward1997104/Wavelet-Generation}, and the class conditioned model from 3DILG\footnote{https://github.com/1zb/3DILG} and 3DShape2VecSet\footnote{https://github.com/1zb/3DShape2VecSet}.


\section{Additional implementation details}
In this section we provide additional implementation details missing from the main paper. 

\vspace{-5pt}
\paragraph{Computing M-SDF representation.}
As described in Section \ref{sec:compute_msdf} and algorithm \ref{alg:mosaicsdf}, the computation of the M-SDF representation for a given shape $\gS$ consists of two stages: initialization and fine-tuning. For both stages we 
require a good estimation of the ground truth SDF $F_\gS$, and for that we use the open-source library of Kaolin-Wisp\footnote{https://github.com/NVIDIAGameWorks/kaolin-wisp}.
To obtain $s$ (equation \ref{e:coverage}), that serves as the initial scale, we sample the surface densely and search for the minimal distance between the dense sampling set and the initialized volume centers.
For the fine-tuning stage we sample for supervision a set of 300K points on the surface, and 200K points near the surface perturbed with Gaussian noise with variance $0.01$. In each fine-tuning step we sample a random batch of 16K points, used to compute the loss in equation \ref{e:rep_loss}. We run the fine-tuning for 1K steps with \textsc{Adam} optimizer \cite{kingma2014adam} and learning rate of $1\mathrm{e}{-4}$.

\vspace{-5pt}
\paragraph{M-SDF representation configuration.}
The number of local grids is chosen to be $n=1024$ as this is the common size of the generated point cloud used in previous point-based diffusion models \cite{nichol2022pointe}. The grid resolution was set to $k=7$, as it is the highest dimension that our transformer architecture can be consistent with. 

\vspace{-5pt}
\paragraph{Other representations configurations.}
In section \ref{s:representation_eval} we compare M-SDF representation to existing popular SDF representations used in 3D generative models.
For the experiment results presented in Figure \ref{fig:rep_eval} we follow the configurations commonly used with these representations in previous works. Specifically, for INR we had 8 hidden layers and changed the width of the hidden layers appropriately to the given parameter budget. As for the 3D grid and triplane we only adjusted the grid's resolution, where the triplane planes have fixed features dimension of 32. 

\vspace{-5pt}
\paragraph{Conditioning tokens.} To complete the architecture description in Section \ref{s:implementation_details}, we add details regarding the conditioning mechanism $\vc$. For the class conditioning generation \ref{sec:class_gen} we use a learned per-class embedding where each class is described using a latent vector of size 128. For a selected class latent we first apply a linear layer projecting it to the transformer dimension, \ie, 1024 before feeding it to the transformer. For the text conditioning generation \ref{s:textcond}, we utilize a pre-trained text model \cite{2020t5} as our textual embedding, result in a token embedding with feature size of 768 and maximum sequence length of 32. We feed these additional 32 tokens to the transformer, after applying a linear layer projecting to 1024.

\input{tables/generation_time}

\paragraph{Generation timing.} In Table \ref{tab:gen_time} we report the time (in seconds) and Number of Function Evaluations (NFE) for generating one sample according to algorithm \ref{alg:sampling}, using different ODE solvers: Midpoint rule, and Dormand–Prince method (DOPRI) \cite{torchdiffeq}. We further report the effect of the different solvers on the quality of the generated shapes using the 1-NNA metric. For this experiment we evaluated our class-conditioning model, on 300 generated samples from the 'airplane' class. Please note that in all of the paper's experiments and evaluations we used the DOPRI as our ODE solver, however as Table \ref{tab:gen_time} indicates using the Midpoint method, with either 25 or 50 steps, results in faster generation and only a mild degradation in quality. Using recent advances in fast sampling of flow models is expected to reduce these times further.


\section{Additional representation evaluations}

\begin{wrapfigure}[11]{r}{0.22\columnwidth}
  \begin{center}\vspace{-22pt}\hspace{-25pt}
    \includegraphics[width=0.29\columnwidth]{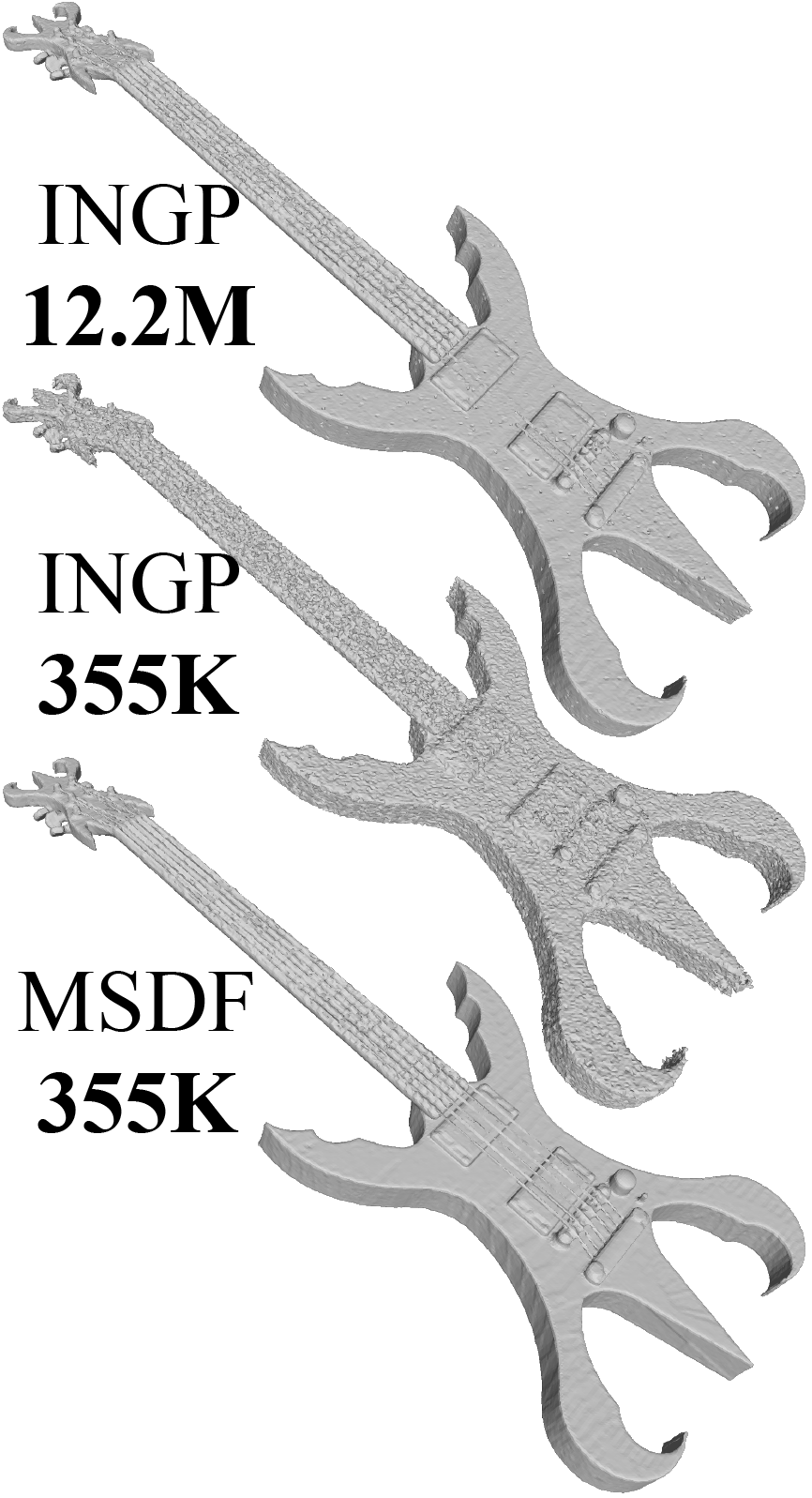}
  \end{center}\vspace{-20pt}
\end{wrapfigure}

\paragraph{Comparison to Instant-NGP representation.} To complement with the M-SDF representation evaluation shown in Figure \ref{fig:msdf}, we further examine the Instant-NGP (INGP) \cite{mueller2022instant} representation power for a fixed parameter budget. Please note that the INGP representation has a more complicated structure, as it consists of a triplet: coarse level grid; ordered set of hash tables; and weights of a small MLP. These lead to a representation incorporating different tensors with various symmetries, which might be possible to work with but was not done in previous 3D generative models and is non-trivial. 
As the inset shows, INGP best performance is achieved when using the INGP's original configuration (12.2M params), which is considerably larger than MSDF (355K params) that still leads to better approximation of the guitar.

\vspace{-10pt}
\begin{wrapfigure}[11]{r}{0.2\columnwidth}
  \begin{center}\vspace{-20pt}\hspace{-23pt}
    \includegraphics[width=0.245\columnwidth]{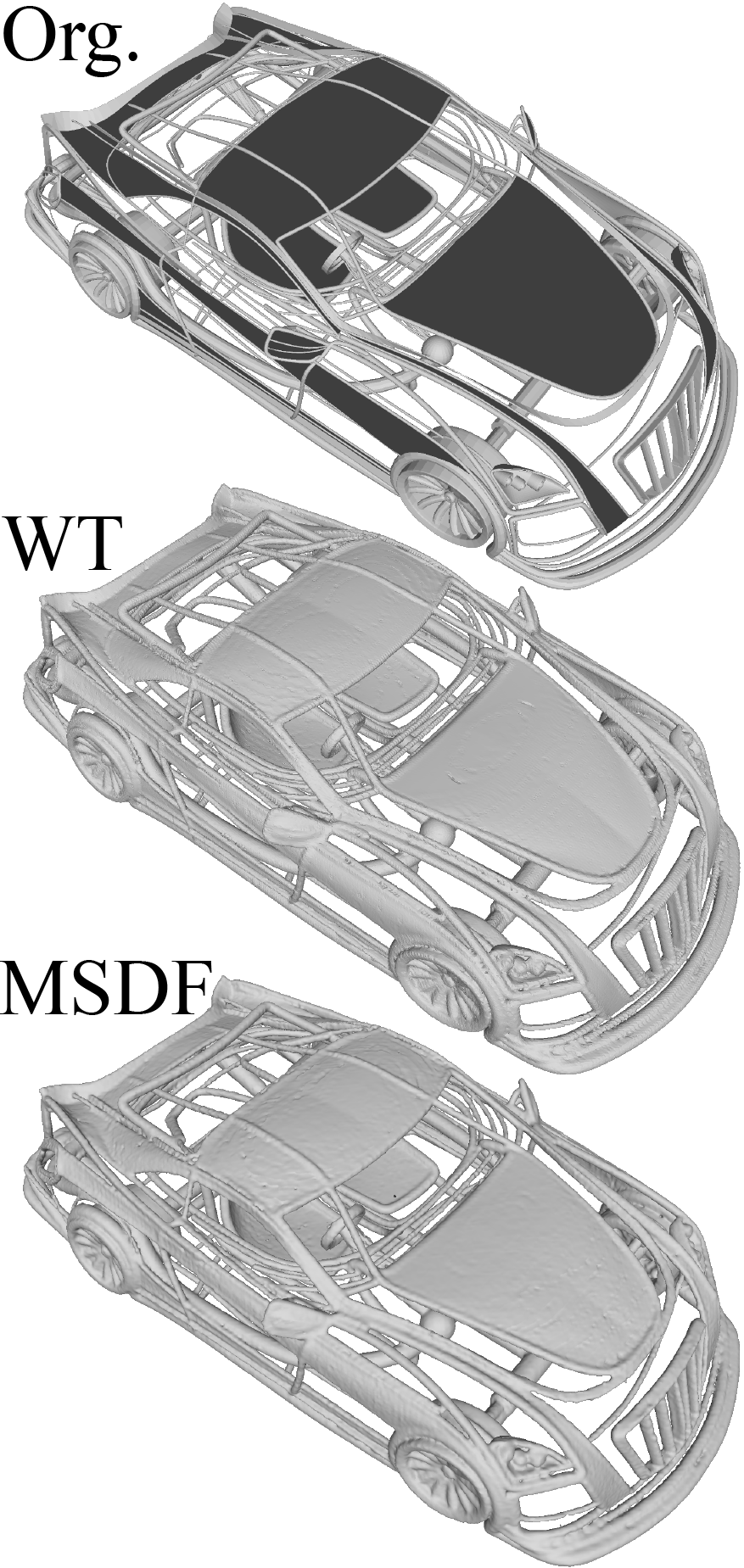}
  \end{center}\vspace{-20pt}
\end{wrapfigure}

\paragraph{The manifold assumption}
Man-made 3D shapes are typically not manifolds, however practically all shapes can be described as manifolds (\eg, with small width). The manifold assumption has its advantages in defining implicit representations and post-processing (extracting a mesh) and therefore widely common assumption in other 3D generative models. 
%
%
In the inset we also add a visualization of a model with thin structures, after its processing to be watertight (WT), and the MSDF representation result. 

\section{Additional results}

\begin{figure}[b!]
\centering
\begin{tabular}{@{\hskip0pt}c@{\hskip3pt}c@{\hskip0pt}c@{\hskip0pt}c@{\hskip0pt}c@{\hskip0pt}c@{\hskip0pt}c@{\hskip0pt}}
\rotatebox{90}{\scriptsize \textbf{\textit{3DILG}}} &
\includegraphics[width=0.09\columnwidth]{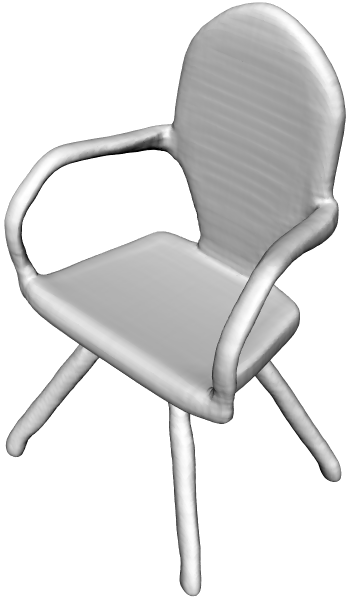} &  
\includegraphics[width=0.10\columnwidth]{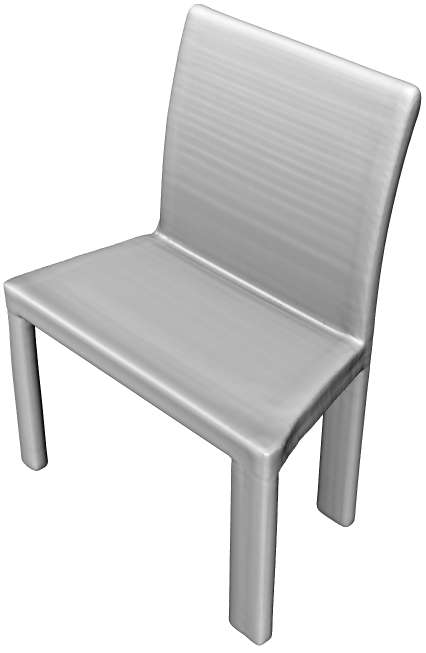} &  
\includegraphics[width=0.18\columnwidth]{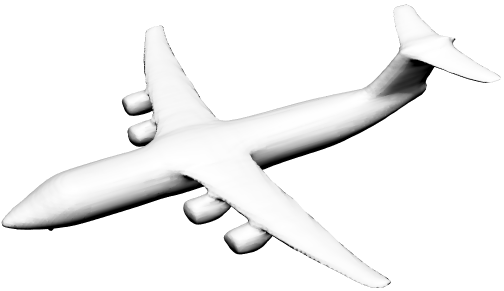} &  
\includegraphics[width=0.17\columnwidth]{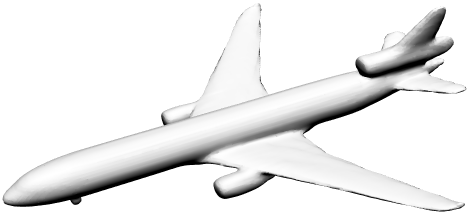} &  
\includegraphics[width=0.16\columnwidth]{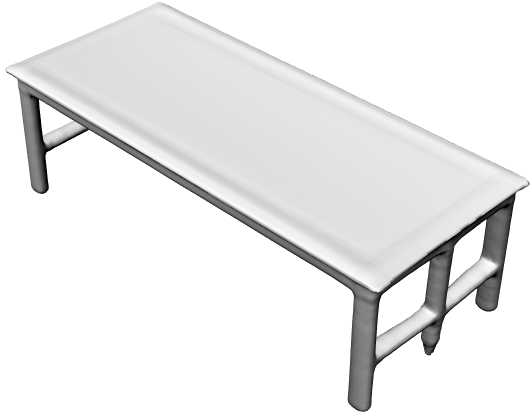} &  
\includegraphics[width=0.16\columnwidth]{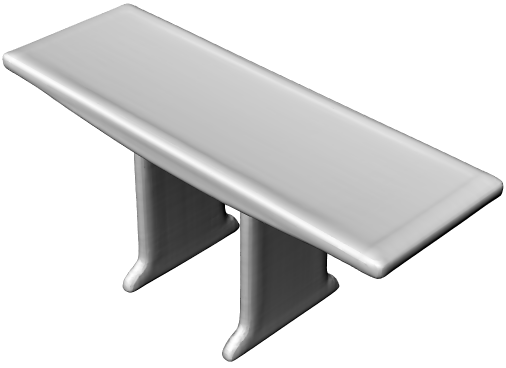} \vspace{-3pt}\\    
\rotatebox{90}{\scriptsize \textbf{\quad \textit{NW}}} &
\includegraphics[width=0.10\columnwidth]{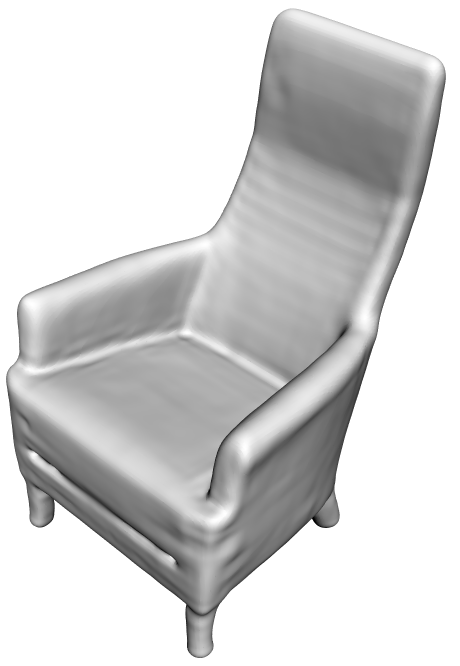} &  
\includegraphics[width=0.08\columnwidth]{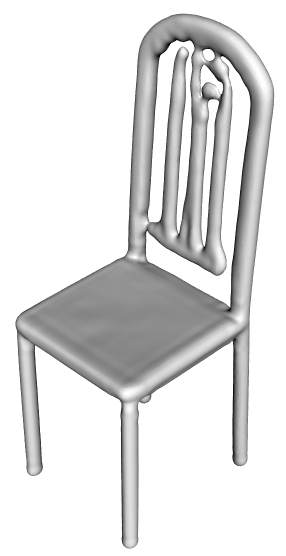} &  
\includegraphics[width=0.18\columnwidth]{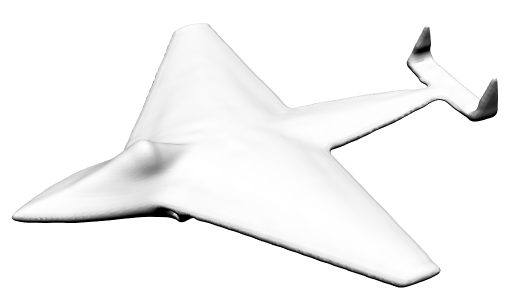} &  
\includegraphics[width=0.18\columnwidth]{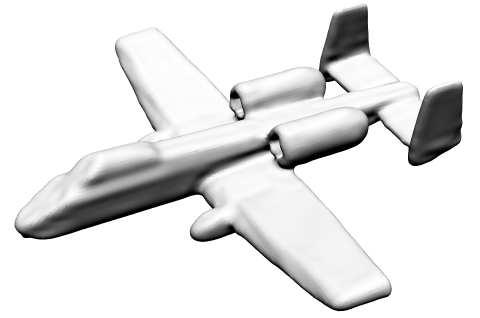} &  
\includegraphics[width=0.15\columnwidth]{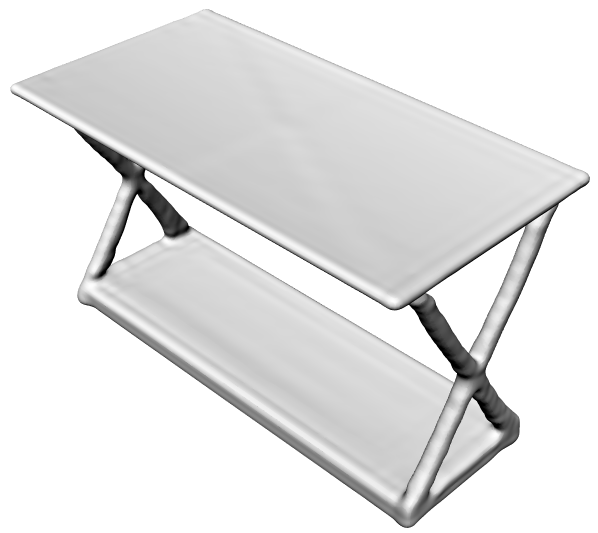} &  
\includegraphics[width=0.16\columnwidth]{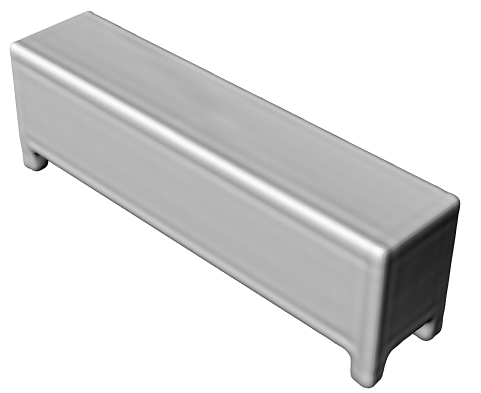} \vspace{-3pt} \\ 
\rotatebox{90}{\scriptsize \textbf{\quad \textit{S2VS}}} &
\includegraphics[width=0.10\columnwidth]{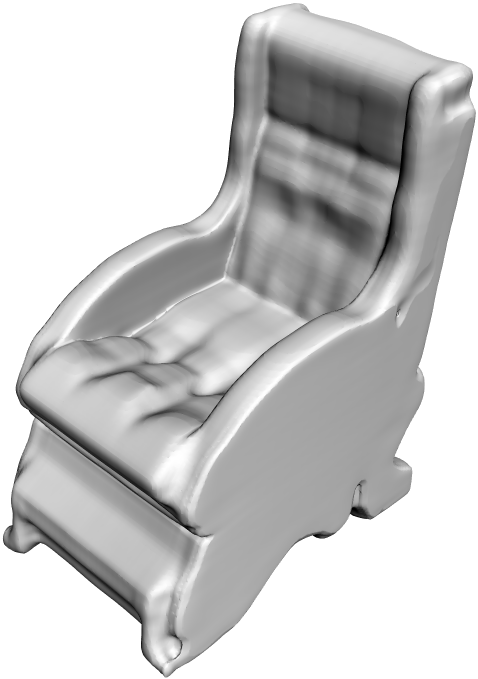} &  
\includegraphics[width=0.11\columnwidth]{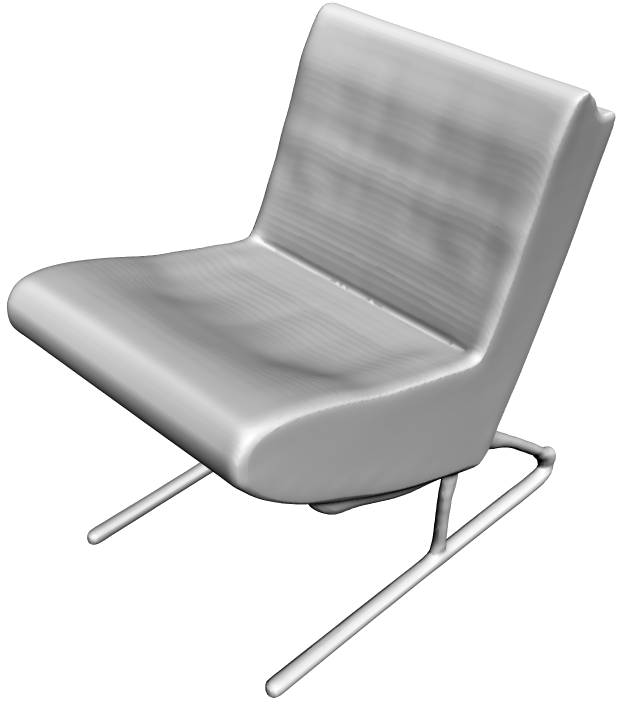} &  
\includegraphics[width=0.18\columnwidth]{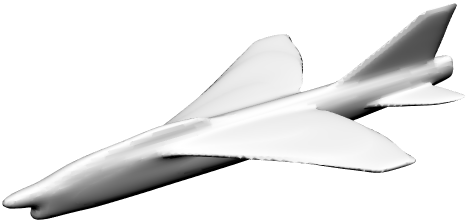} &  
\includegraphics[width=0.18\columnwidth]{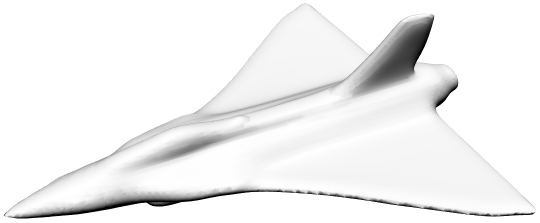} &  
\includegraphics[width=0.13\columnwidth]{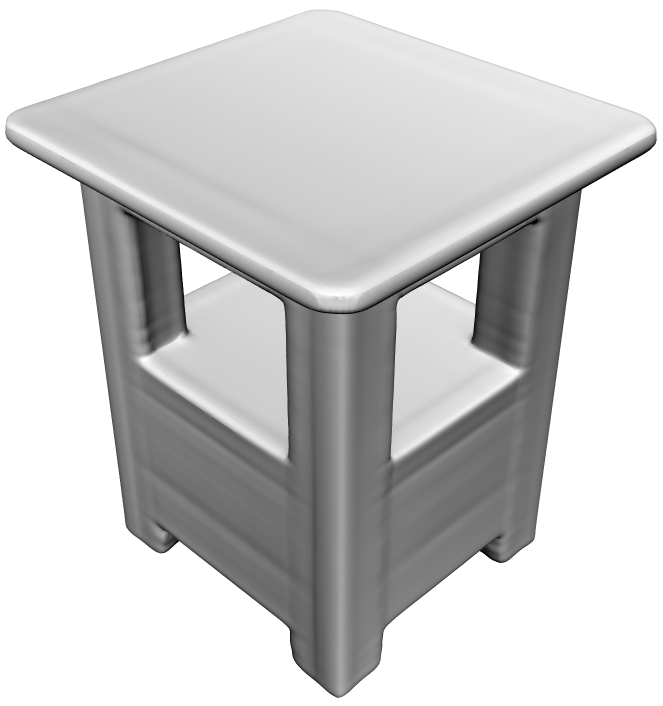} &  
\includegraphics[width=0.14\columnwidth]{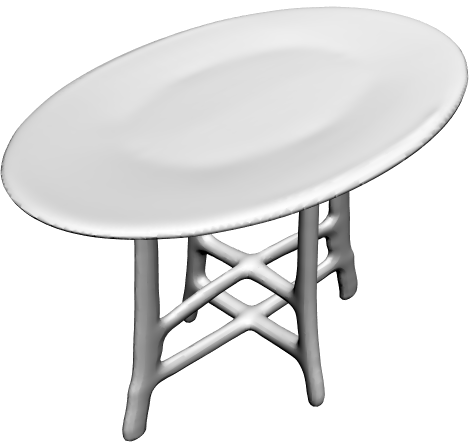} \vspace{-3pt} \\ 
\rotatebox{90}{\scriptsize \textbf{\textit{\quad Ours}}} &
\includegraphics[width=0.09\columnwidth]{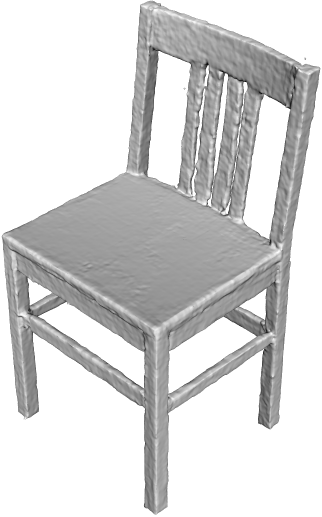} &  
\includegraphics[width=0.08\columnwidth]{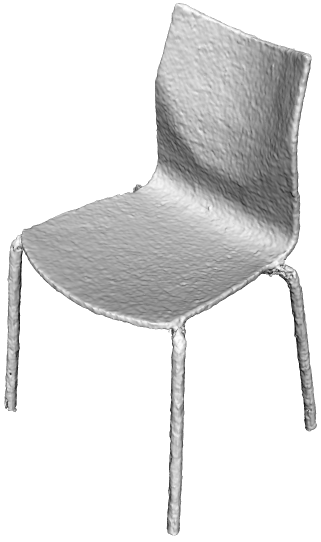} &  
\includegraphics[width=0.17\columnwidth]{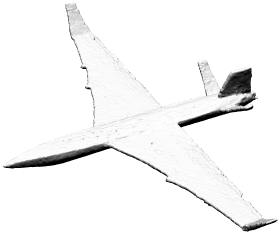} &  
\includegraphics[width=0.18\columnwidth]{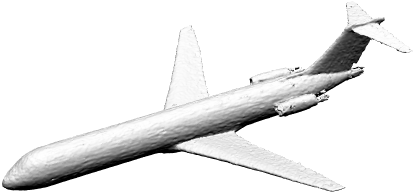} &  
\includegraphics[width=0.16\columnwidth]{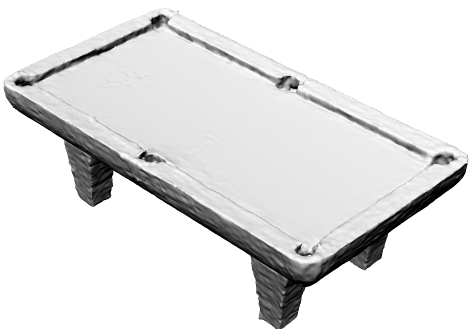} &  
\includegraphics[width=0.16\columnwidth]{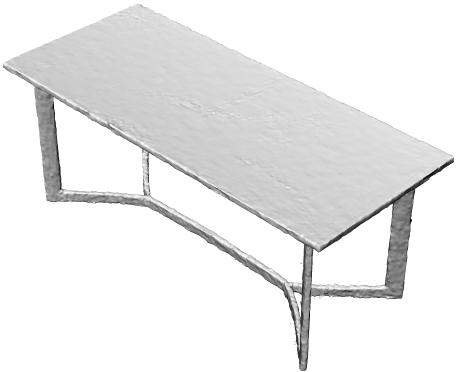} \vspace{3pt} \\
\hdashline \vspace{-8pt}\\
\rotatebox{90}{\scriptsize \textbf{\quad \textit{3DILG}}} &
\includegraphics[width=0.09\columnwidth]{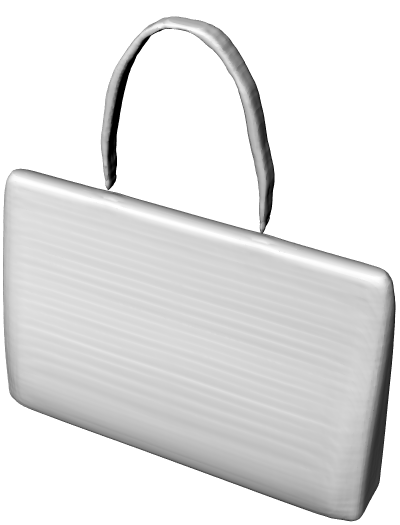} &  
\includegraphics[width=0.12\columnwidth]{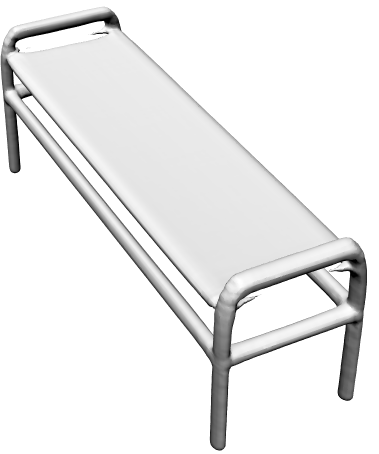} &  
\includegraphics[width=0.12\columnwidth]{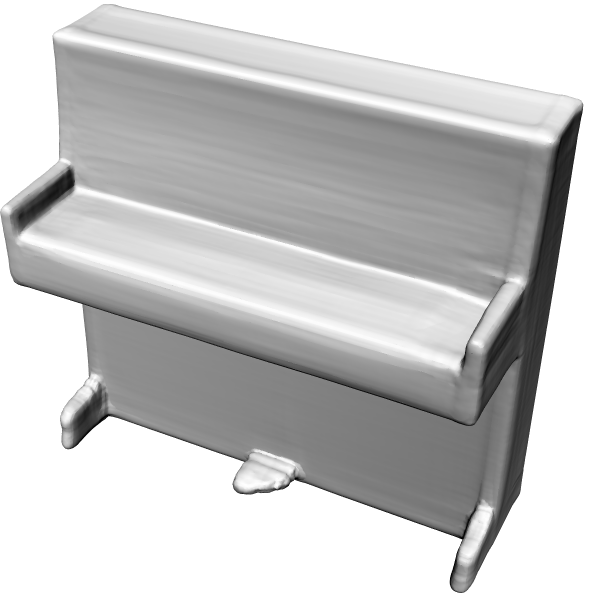} &  
\includegraphics[width=0.18\columnwidth]{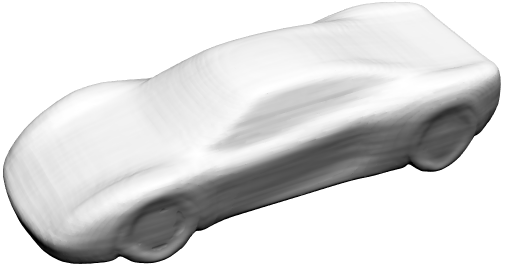} &  
\includegraphics[width=0.14\columnwidth]{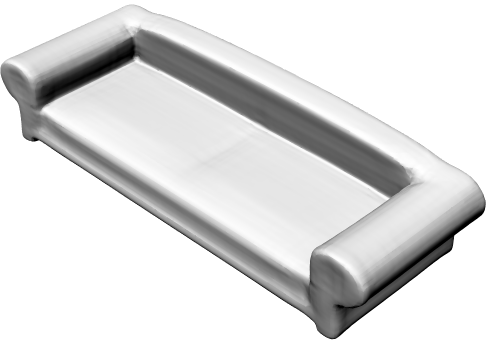} &  
\includegraphics[width=0.055\columnwidth]{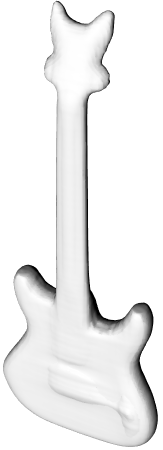} \vspace{-3pt} \\ 
\rotatebox{90}{\scriptsize \textbf{\quad \textit{S2VS}}} &
\includegraphics[width=0.09\columnwidth]{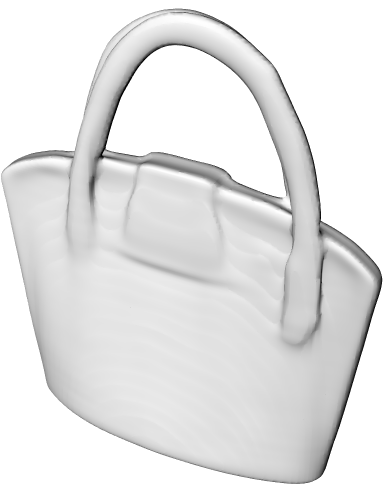} &  
\includegraphics[width=0.12\columnwidth]{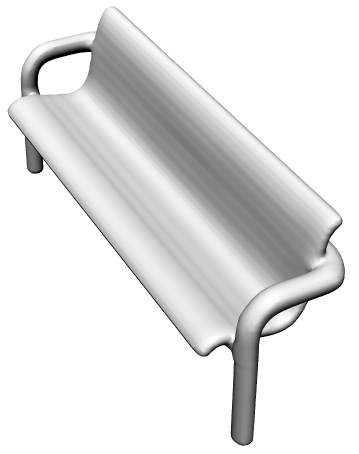} &  
\includegraphics[width=0.12\columnwidth]{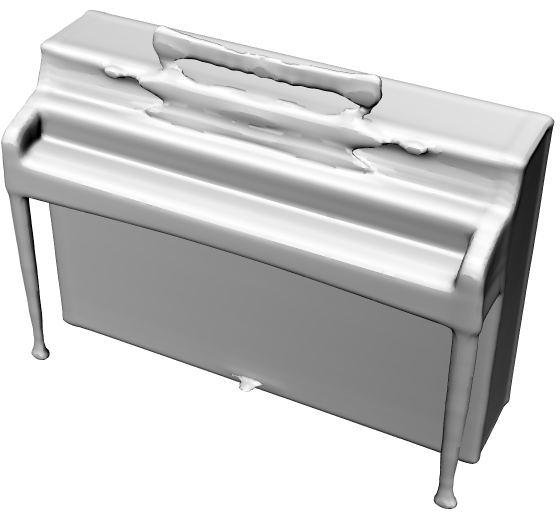} &  
\includegraphics[width=0.18\columnwidth]{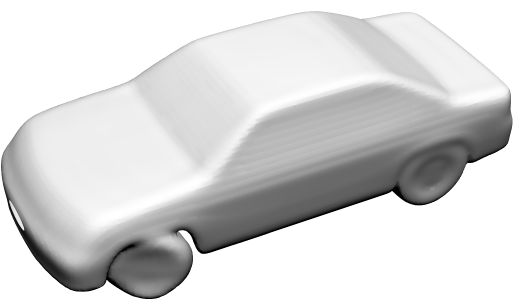} &  
\includegraphics[width=0.14\columnwidth]{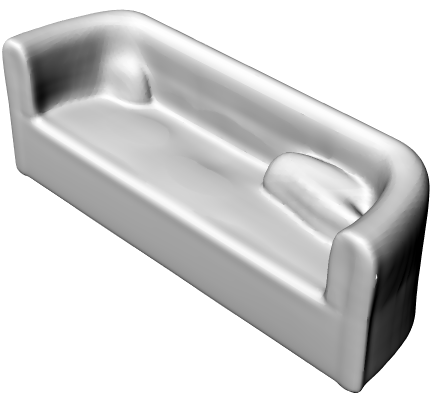} &  
\includegraphics[width=0.06\columnwidth]{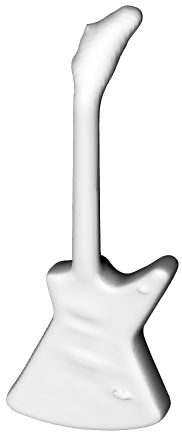}  \vspace{-3pt} \\ 
\rotatebox{90}{\scriptsize \textbf{\textit{\quad Ours}}} &
\includegraphics[width=0.09\columnwidth]{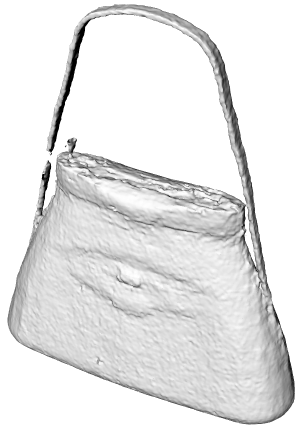} &  
\includegraphics[width=0.14\columnwidth]{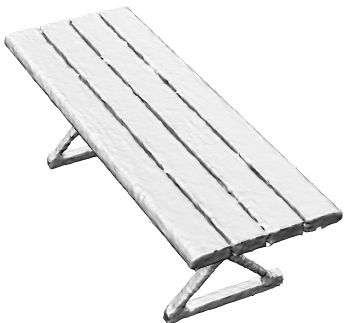} & 
\includegraphics[width=0.13\columnwidth]{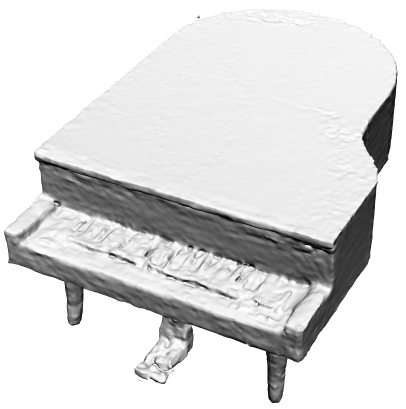}  & 
\includegraphics[width=0.18\columnwidth]{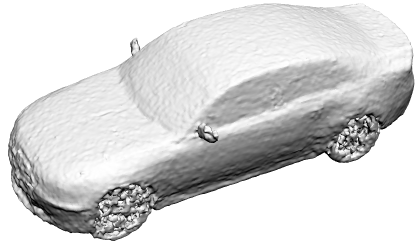} &  
\includegraphics[width=0.16\columnwidth]{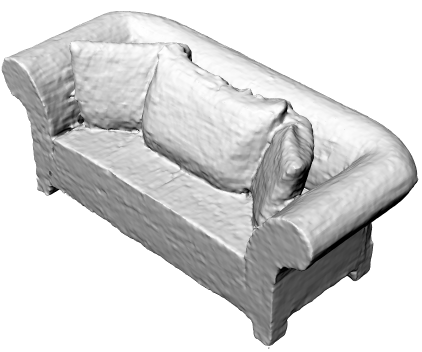} &   
\includegraphics[width=0.055\columnwidth]
{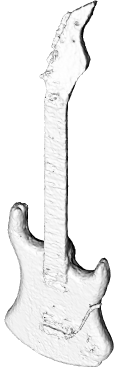} \vspace{-5pt}
\end{tabular}
    \vspace{-3pt}
     \caption{Class conditioned generation of 3D shapes compared to relevant baselines.}\label{fig:shapenet_compare_supp}
\end{figure}

\vspace{-5pt}
\paragraph{Class conditioning generation.} In Figure \ref{fig:shapenet_compare_supp} we show additional qualitative comparison of the class-conditioned generation compared to the relevant baselines. On top we show the common classes across all baselines. Below the dashed line we further present other classes in a comparison to 3DILG\cite{zhang2022dilg} and S2VS\cite{zhang20233dshape2vecset}, which trained a class conditioning model similarly to us. Note that M-SDF generation are overall sharper with more details, while baselines tend to over smooth. 

\begin{figure}[t!]
\centering
\begin{tabular}{@{\hskip0pt}c@{\hskip4pt}c@{\hskip4pt}c@{\hskip4pt}c@{\hskip4pt}c@{\hskip0pt}}
\includegraphics[width=0.17\columnwidth]{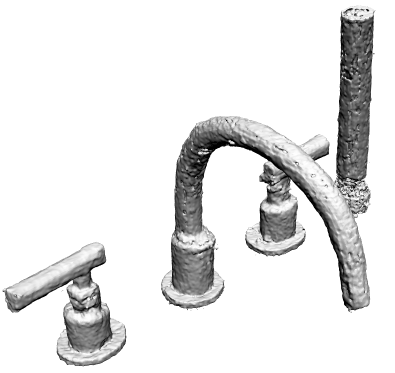} &  
\includegraphics[width=0.17\columnwidth]{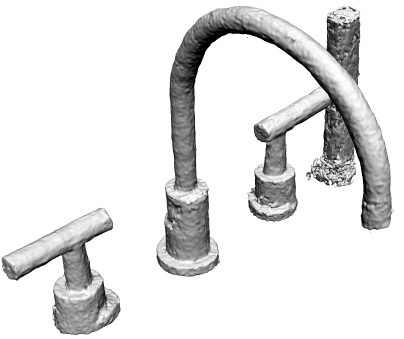} &  
\includegraphics[width=0.17\columnwidth]{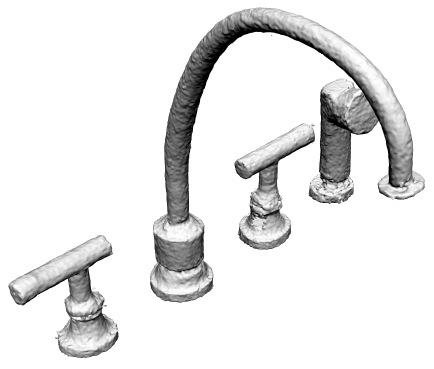} &  
\includegraphics[width=0.17\columnwidth]{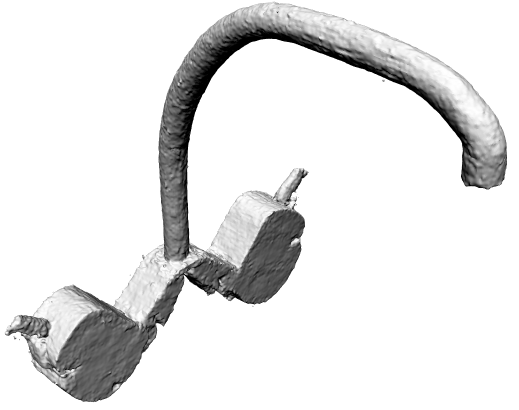} &  
\includegraphics[width=0.17\columnwidth]{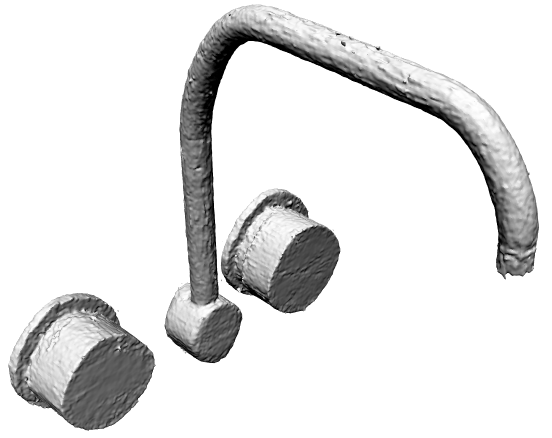} 
\vspace{-4pt} \\
\includegraphics[width=0.12\columnwidth]{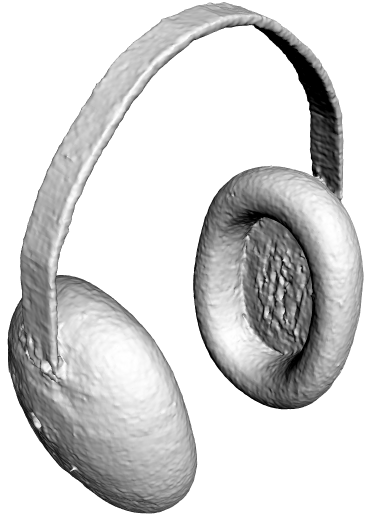} &  
\includegraphics[width=0.12\columnwidth]{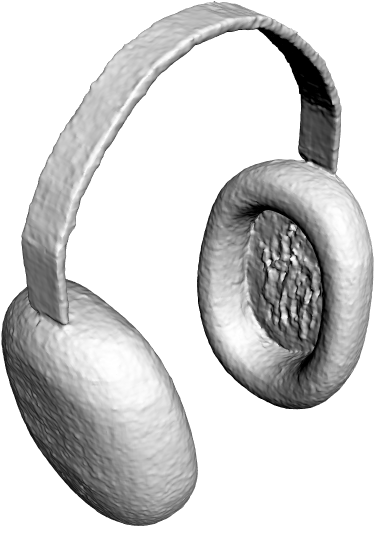} &  
\includegraphics[width=0.14\columnwidth]{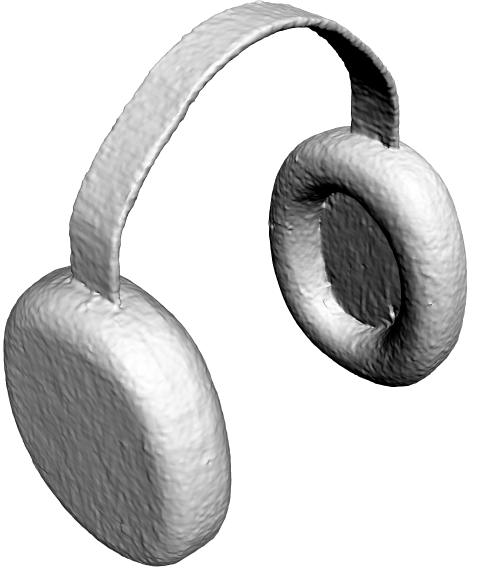} &  
\includegraphics[width=0.14\columnwidth]{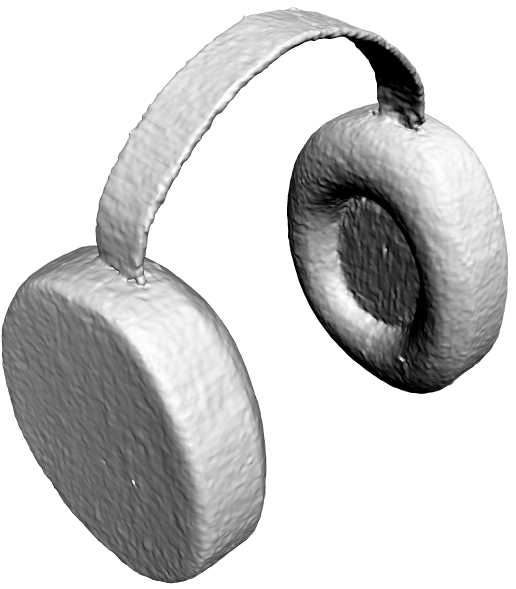} &  
\includegraphics[width=0.14\columnwidth]{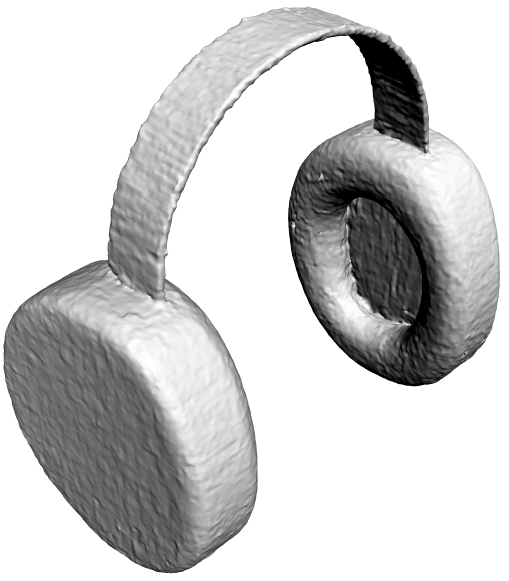}
\vspace{-4pt} \\
\includegraphics[width=0.17\columnwidth]{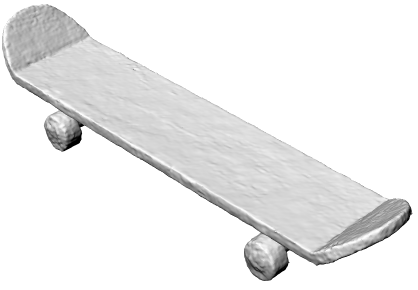} &  
\includegraphics[width=0.17\columnwidth]{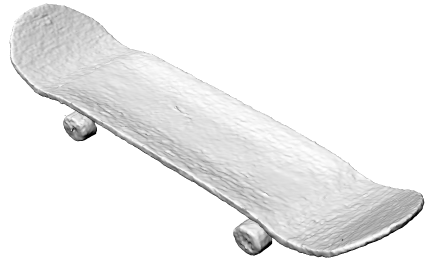} &  
\includegraphics[width=0.17\columnwidth]{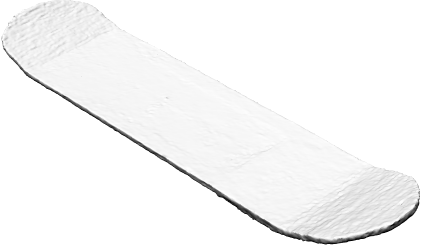} &  
\includegraphics[width=0.17\columnwidth]{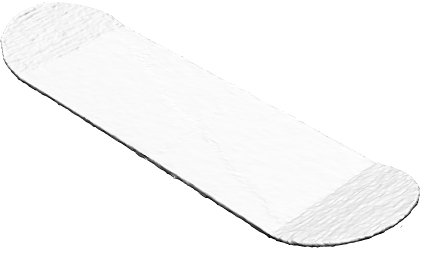} &  
\includegraphics[width=0.17\columnwidth]{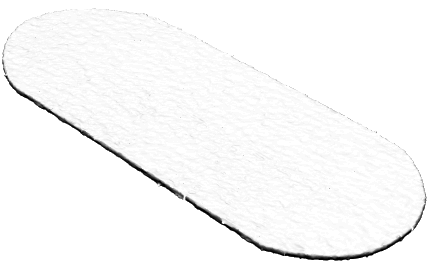} 
\vspace{0pt} \\
\hdashline \vspace{-12pt} \\ 
\multicolumn{5}{c}{\emph{\scriptsize{"Ferris wheel"}}} \vspace{-2pt} \\
\includegraphics[width=0.14\columnwidth]{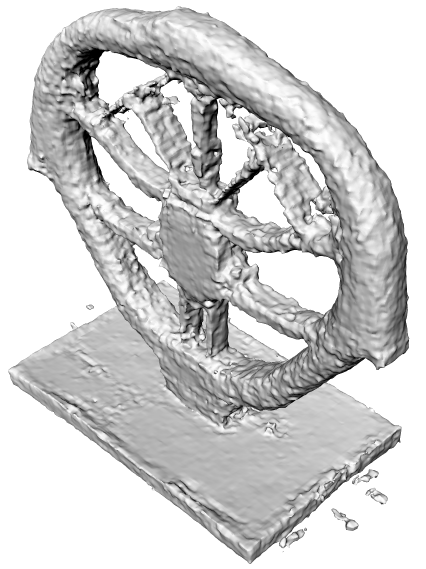} &  
\includegraphics[width=0.13\columnwidth]{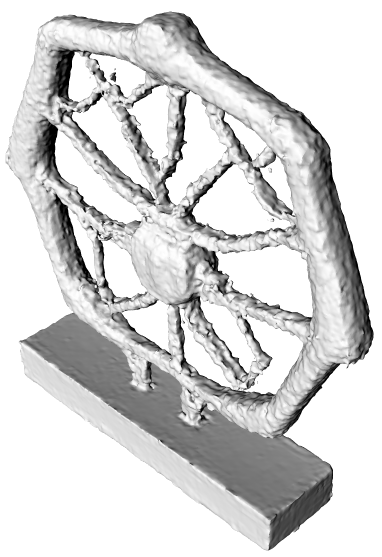} &  
\includegraphics[width=0.13\columnwidth]{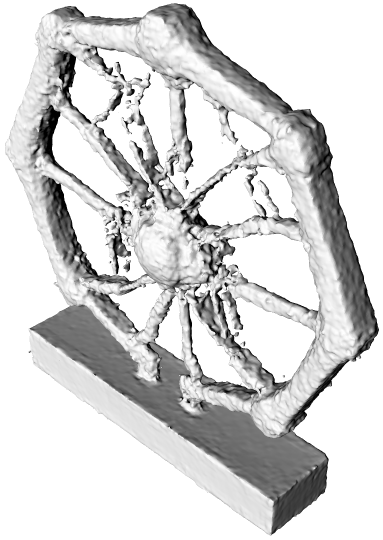} &  
\includegraphics[width=0.13\columnwidth]{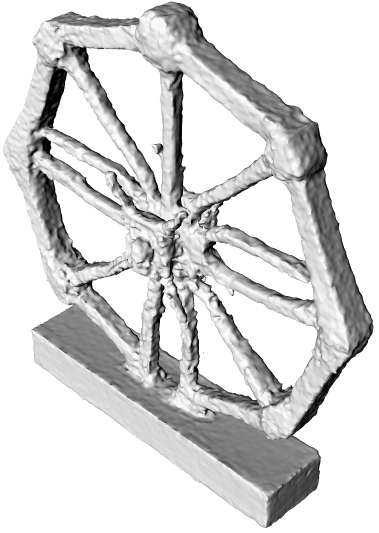} &  
\includegraphics[width=0.13\columnwidth]{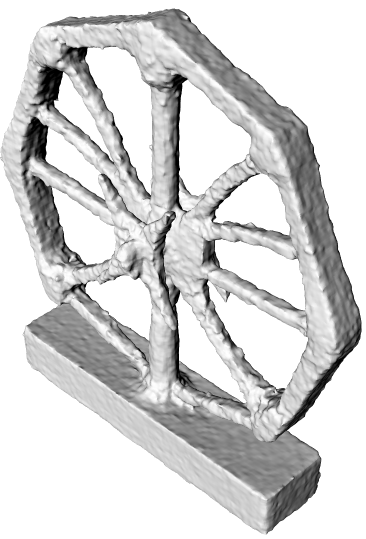} 
\vspace{-5pt} \\ 
\multicolumn{5}{c}{\emph{\scriptsize{"A teddy bear"}}} \vspace{-3pt} \\
\includegraphics[width=0.14\columnwidth]{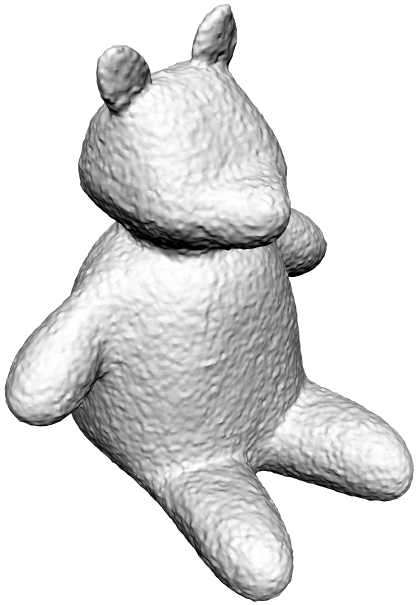} &  
\includegraphics[width=0.14\columnwidth]{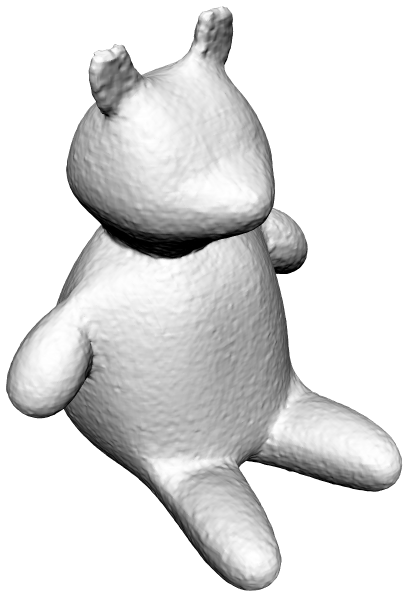} &  
\includegraphics[width=0.14\columnwidth]{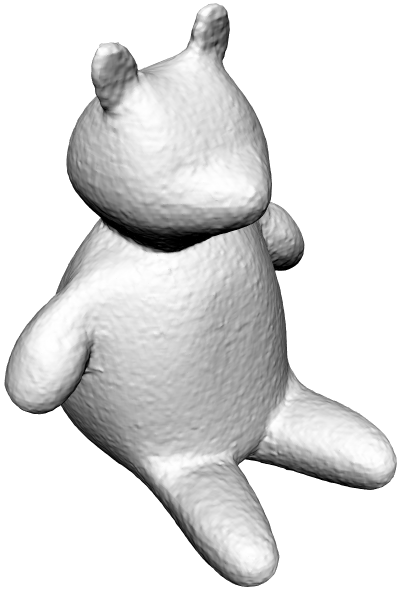} &  
\includegraphics[width=0.14\columnwidth]{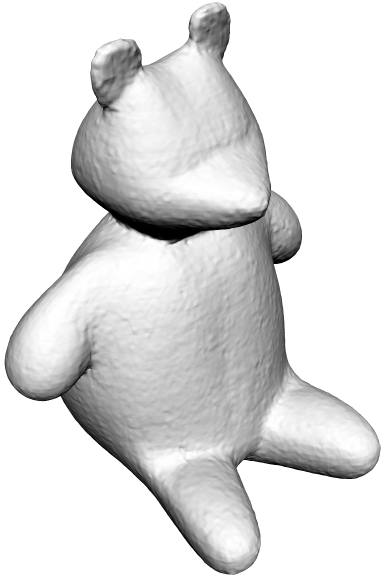} &  
\includegraphics[width=0.14\columnwidth]{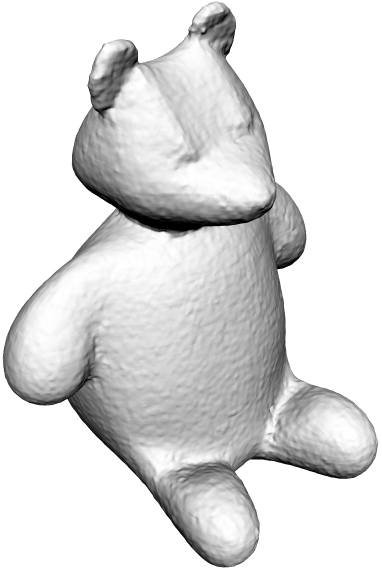}
\vspace{-4pt} \\
\multicolumn{5}{c}{\emph{\scriptsize{"A beautiful earring with diamonds"}}} \vspace{-3pt} \\
\includegraphics[width=0.09\columnwidth]{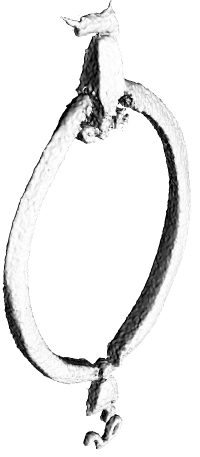} &  
\includegraphics[width=0.10\columnwidth]{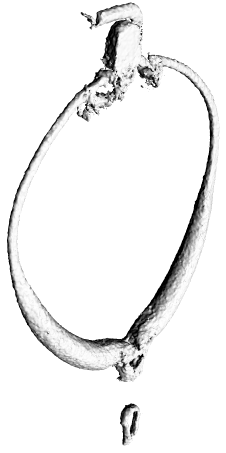} &  
\includegraphics[width=0.10\columnwidth]{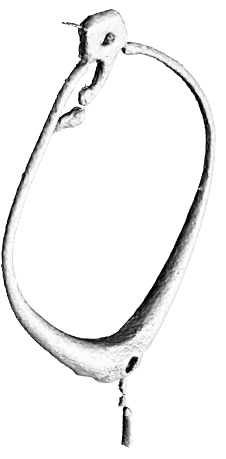} &  
\includegraphics[width=0.10\columnwidth]{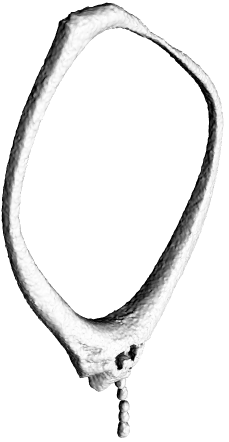} &  
\includegraphics[width=0.10\columnwidth]{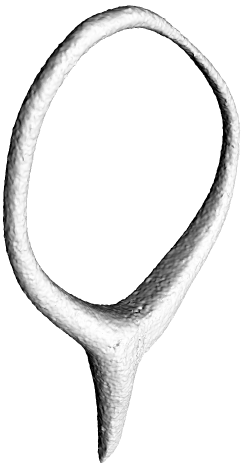}
\\
$\omega=0$ & $\omega=1$ & $\omega=2$ & $\omega=5$ & $\omega=10$
\vspace{-10pt}
\end{tabular}
     \caption{
     \footnotesize Ablation of guidance scale $\omega$ use in sampling our class-conditioned model (top) and our text-conditioned
     model (bottom).
     }\label{fig:guidence_ablation}
\end{figure}

\vspace{-8pt}
\paragraph{Guidance scale ablation.} We perform an ablation study regarding the guidance scale $\omega$ we use in the sampling algorithm \ref{alg:sampling}. Figure \ref{fig:guidence_ablation} depicts the generation samples using different guidance scales, with both our class-conditioning model (top) and the text-conditioning model (bottom).
We further provide quantitative comparison in Table \ref{tab:guidence_ablation}, when sampling our class conditioning model using various guiding scales. Here, we perform similar evaluation to the class conditioning evaluation in Table \ref{tab:shapenet_cond}, and report metrics for the 5 largest classes in the ShapeNetCore-V2 (3D Warehouse) \cite{shapenet2015} dataset. As somewhat expected, the $\omega=0$ performs best when comparing shape \emph{distributions}, however qualitatively, taking a higher $\omega$ tends to result in a more "common" or "average" shape. In the main paper we therefore opted $\omega=0$ for the class-conditional shape generation, and $\omega=5$ for the text-conditioned shape generation.

\input{tables/guidence_ablation}

\begin{figure}[h]
    \centering
    \includegraphics[width=1.\linewidth]{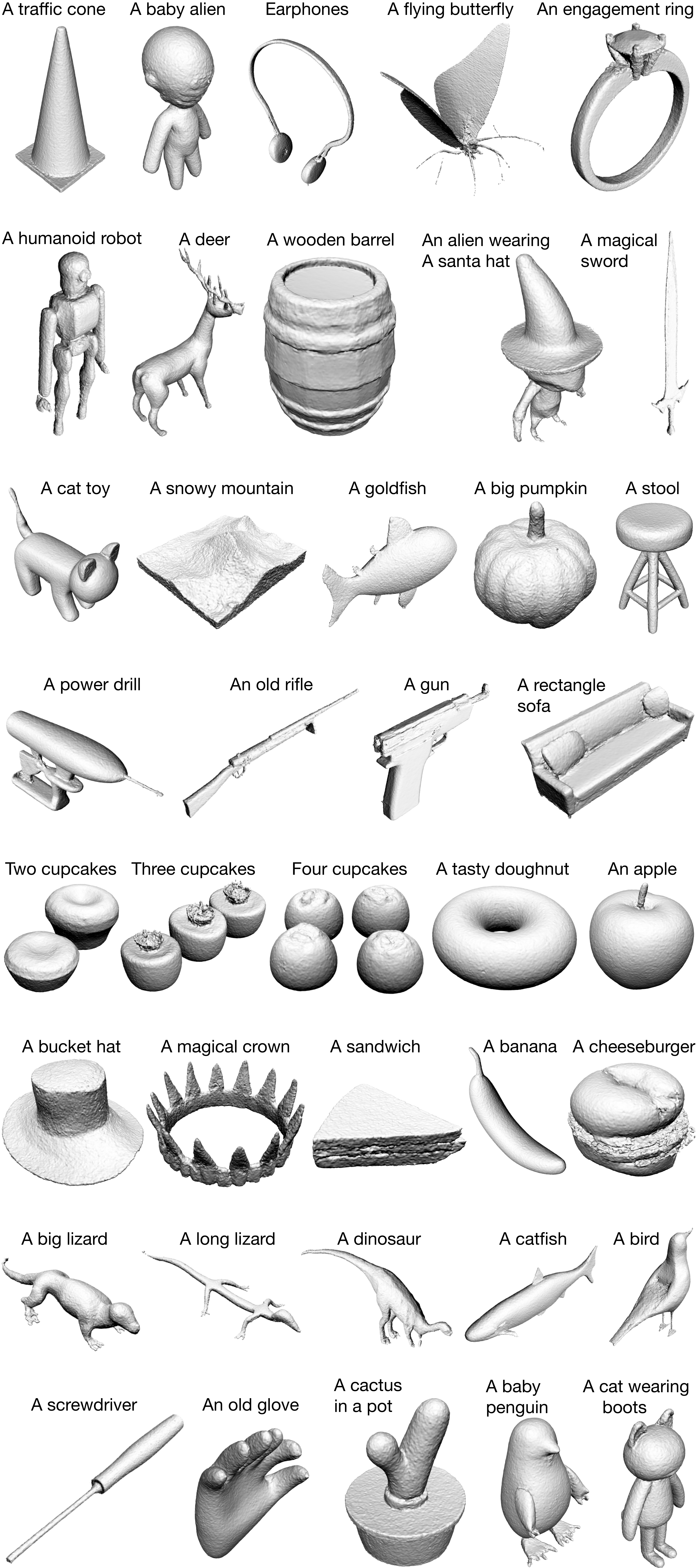}
    \caption{Additional text-to-3D samples from a Flow Matching model trained on M-SDF representations of 600K pairs of shapes and text.}
    \vspace{5pt}
    \label{fig:text_to_3d_generation_supp}
\end{figure}

\paragraph{Text-to-3D generation.} In Figure \ref{fig:text_to_3d_generation_supp} we show additional generated shapes from our text-conditioned model.

%% file: tables/generation_time.tex
\begin{wraptable}[7]{r}{0.545\linewidth}
    \centering
    \resizebox{\linewidth}{!}{
\begin{tabular}{lcccc} 
     \toprule
       & NFE & Time & \multicolumn{2}{c}{1-NNA ($\downarrow$,\%)} \\
       & & (sec.) & \small{CD}  & \small{EMD}
      \\ \midrule 
     \textbf{\textit{Midpoint-25}} \ 
     &
    50.00
    &
    6.13
    &
    59.16
    &
    67.57
    \\
     \textbf{\textit{Midpoint-50}} \ 
     &
    100.00
    &
    12.19
    &
    61.88
    &
    69.06
    \\
      \textbf{\textit{DOPRI}} \ 
     &
    138.46
    &
    16.97
    &
    57.67
    &
    64.85
    \\
     \bottomrule 
    \end{tabular} 
    }
    \vspace{-5pt}
    \caption{Generation complexity and quality for different ODE solvers.
    }\label{tab:gen_time}
\end{wraptable}

%% file: tables/guidence_ablation.tex
\renewcommand{\arraystretch}{1.0} 
\begin{table}[h]
    \centering
    \resizebox{0.98\linewidth}{!}{
    \begin{tabular}{lcccccccc} 
     \toprule
       &  FPD ($\downarrow$) & KPD ($\downarrow$) &  \multicolumn{2}{c}{COV ($\uparrow$,\%)} & \multicolumn{2}{c}{MMD ($\downarrow$)} & \multicolumn{2}{c}{1-NNA ($\downarrow$,\%)}\\
       &   &  &  \small{CD}  & \small{EMD} & \small{CD} & \small{EMD} & \small{CD} & \small{EMD} \\
       \bottomrule
 airplane & & & & & & & &      \\ \midrule
\textbf{\textit{Ours $\omega=0$}} & \cellcolor{lightblue} {0.37} & \cellcolor{lightblue} {0.37} & \cellcolor{lightblue} {50.99} & \cellcolor{lightblue} {48.02} & \cellcolor{lightblue} {3.46} & \cellcolor{lightblue} {3.71} & \cellcolor{lightblue} {57.67} & \cellcolor{lightblue} {64.85} \\
\textbf{\textit{Ours $\omega=1$}} & \cellcolor{lightgray} {0.71} & \cellcolor{lightgray} {0.75} & \cellcolor{lightgray} {44.06} & 40.59 & \cellcolor{lightgray} {4.03} & 3.92 & \cellcolor{lightgray} {65.35} & 74.75 \\
\textbf{\textit{Ours $\omega=2$}} & 0.80 & 0.79 & 37.13 & \cellcolor{lightgray} {41.09} & 4.65 & \cellcolor{lightgray} {3.84} & 70.54 & \cellcolor{lightgray} {73.27} \\
\textbf{\textit{Ours $\omega=5$}} & 1.10 & 1.19 & 36.63 & 35.64 & 4.90 & 4.14 & 74.50 & 78.22 \\
\textbf{\textit{Ours $\omega=10$}} & 1.97 & 2.90 & 27.72 & 27.72 & 6.25 & 4.54 & 86.14 & 81.93 \\
       \bottomrule
 car & & & & & & & &      \\ \midrule
\textbf{\textit{Ours $\omega=0$}} & \cellcolor{lightblue} {0.45} & \cellcolor{lightblue} {0.47} & \cellcolor{lightblue} {42.86} & \cellcolor{lightblue} {45.14} & \cellcolor{lightblue} {2.75} & \cellcolor{lightblue} {2.78} & \cellcolor{lightblue} {65.71} & \cellcolor{lightblue} {70.00} \\
\textbf{\textit{Ours $\omega=1$}} & \cellcolor{lightgray} {0.85} & \cellcolor{lightgray} {1.00} & \cellcolor{lightgray} {32.00} & \cellcolor{lightgray} {37.71} & \cellcolor{lightgray} {3.14} & 2.87 & 77.14 & \cellcolor{lightgray} {72.29} \\
\textbf{\textit{Ours $\omega=2$}} & 0.96 & 1.15 & 29.71 & 35.43 & 3.24 & \cellcolor{lightgray} {2.85} & \cellcolor{lightgray} {75.14} & 72.29 \\
\textbf{\textit{Ours $\omega=5$}} & 1.08 & 1.28 & 28.57 & 34.86 & 3.41 & 3.02 & 80.86 & 74.86 \\
\textbf{\textit{Ours $\omega=10$}} & 1.39 & 2.11 & 24.00 & 28.57 & 3.53 & 3.18 & 85.71 & 83.71 \\
       \bottomrule
 chair & & & & & & & &      \\ \midrule
\textbf{\textit{Ours $\omega=0$}} & \cellcolor{lightblue} {0.51} & \cellcolor{lightblue} {0.20} & \cellcolor{lightblue} {45.86} & \cellcolor{lightblue} {51.48} & \cellcolor{lightblue} {16.08} & \cellcolor{lightblue} {9.17} & \cellcolor{lightblue} {55.92} & \cellcolor{lightblue} {55.47} \\
\textbf{\textit{Ours $\omega=1$}} & \cellcolor{lightgray} {0.78} & \cellcolor{lightgray} {0.64} & \cellcolor{lightgray} {44.08} & \cellcolor{lightgray} {42.60} & \cellcolor{lightgray} {18.70} & \cellcolor{lightgray} {10.39} & \cellcolor{lightgray} {56.07} & \cellcolor{lightgray} {64.79} \\
\textbf{\textit{Ours $\omega=2$}} & 0.94 & 0.85 & 38.46 & 42.60 & 20.16 & 10.59 & 66.27 & 70.56 \\
\textbf{\textit{Ours $\omega=5$}} & 1.34 & 1.42 & 30.77 & 33.43 & 22.43 & 11.13 & 74.26 & 74.26 \\
\textbf{\textit{Ours $\omega=10$}} & 1.67 & 1.92 & 31.07 & 30.47 & 22.51 & 11.62 & 76.18 & 77.96 \\
       \bottomrule
 sofa & & & & & & & &      \\ \midrule
\textbf{\textit{Ours $\omega=0$}} & \cellcolor{lightblue} {0.64} & \cellcolor{lightblue} {0.65} & \cellcolor{lightblue} {44.94} & \cellcolor{lightblue} {50.63} & \cellcolor{lightblue} {11.21} & \cellcolor{lightblue} {7.21} & \cellcolor{lightblue} {59.49} & \cellcolor{lightblue} {58.23} \\
\textbf{\textit{Ours $\omega=1$}} & \cellcolor{lightgray} {1.31} & \cellcolor{lightgray} {1.64} & \cellcolor{lightgray} {36.08} & \cellcolor{lightgray} {36.71} & \cellcolor{lightgray} {15.31} & \cellcolor{lightgray} {8.08} & \cellcolor{lightgray} {69.30} & \cellcolor{lightgray} {63.92} \\
\textbf{\textit{Ours $\omega=2$}} & 1.68 & 2.25 & 27.85 & 34.81 & 17.50 & 8.44 & 80.70 & 72.78 \\
\textbf{\textit{Ours $\omega=5$}} & 2.51 & 4.02 & 22.15 & 31.65 & 20.03 & 8.89 & 87.66 & 79.43 \\
\textbf{\textit{Ours $\omega=10$}} & 3.48 & 6.57 & 17.72 & 25.32 & 21.07 & 9.86 & 89.56 & 80.70 \\
       \bottomrule
 table & & & & & & & &      \\ \midrule
\textbf{\textit{Ours $\omega=0$}} & \cellcolor{lightblue} {0.49} & \cellcolor{lightblue} {0.18} & \cellcolor{lightblue} {52.26} & \cellcolor{lightblue} {55.58} & \cellcolor{lightblue} {13.10} & \cellcolor{lightblue} {7.60} & \cellcolor{lightblue} {52.14} & \cellcolor{lightblue} {51.54} \\
\textbf{\textit{Ours $\omega=1$}} & \cellcolor{lightgray} {1.26} & \cellcolor{lightgray} {1.43} & \cellcolor{lightgray} {39.90} & \cellcolor{lightgray} {43.47} & \cellcolor{lightgray} {15.16} & \cellcolor{lightgray} {8.47} & \cellcolor{lightgray} {65.20} & \cellcolor{lightgray} {63.06} \\
\textbf{\textit{Ours $\omega=2$}} & 1.97 & 2.55 & 32.30 & 31.12 & 18.63 & 9.81 & 73.40 & 73.99 \\
\textbf{\textit{Ours $\omega=5$}} & 3.08 & 4.55 & 17.81 & 18.05 & 36.52 & 14.99 & 89.31 & 88.95 \\
\textbf{\textit{Ours $\omega=10$}} & 4.34 & 8.34 & 13.54 & 14.25 & 54.02 & 19.28 & 95.72 & 94.42 \\
       \bottomrule
    \end{tabular}
    }\vspace{-7pt}
        \caption{
        \footnotesize Ablation study on the Classifier Free Guidance (CFG) scale used for sampling $\omega$. KPD and MMD-CD multiplied by $10^3$, MMD-EMD by $10^2$.
}\label{tab:guidence_ablation}
\end{table}